\documentclass[letterpaper,twocolumn,10pt]{article}
\usepackage{usenix-2020-09}

\usepackage{package}
\usepackage{macros}
\usepackage{marginnote}
\usepackage[available,functional,reproduced]{usenixbadges}

\begin{document}
\date{}
\title{CoVA: Exploiting Compressed-Domain Analysis to Accelerate Video Analytics\vspace{1ex}}

\author{
{\rm Jinwoo Hwang}\\
KAIST
\and
{\rm Minsu Kim}\\
KAIST
\and
{\rm Daeun Kim}\\
KAIST
\and
{\rm Seungho Nam}\\
KAIST
\and
{\rm Yoonsung Kim}\\
KAIST
\and
{\rm Dohee Kim}\\
KAIST
\and
{\rm Hardik Sharma}\\
Google
\and
{\rm Jongse Park}\\
KAIST
}

\maketitle

\begin{abstract}

Modern retrospective analytics systems leverage cascade architecture to mitigate bottleneck for computing deep neural networks (DNNs).
However, the existing cascades suffer from two limitations: (1) decoding bottleneck is either neglected or circumvented, paying significant compute and storage cost for pre-processing; and (2) the systems are specialized for temporal queries and lack spatial query support. 
This paper presents \cova, a novel cascade architecture that splits the cascade computation between compressed domain and pixel domain to address the decoding bottleneck, supporting both temporal and spatial queries.
\cova cascades analysis into three major stages where the first two stages are performed in compressed domain, while the last one in pixel domain.
First, \cova detects occurrences of moving objects (called \emph{blobs}) over a set of compressed frames (called \emph{tracks}). 
Then, using the track results, \cova prudently selects a minimal set of frames to obtain the label information and only decode them to compute the full DNNs, alleviating the decoding bottleneck.
Lastly, \cova associates tracks with labels to produce the final analysis results on which users can process both temporal and spatial queries. 
Our experiments demonstrate that \cova offers 4.8$\times$ throughput improvement over modern cascade systems, while imposing modest accuracy loss.

\end{abstract}

 \section{Introduction}
\label{sec:intro}

Every day, a massive corpus of video data is produced, which is only growing (9.4 exabytes per day, as of 2021~\cite{cisco-forecast}).
Extracting insights and actionable semantics from the captured video can enable a variety of applications in healthcare, smart cities, security, customer behavior analysis, etc.
Prior works~\cite{noscope, tahoma, vstore, probabilistic-predicates, smol, video-monitoring-queries} have built \emph{retrospective analytics systems} that allow analysts to interactively query over a large corpus of accumulated video data stored in disk.

Modern retrospective analytics heavily rely on deep neural networks (DNNs).
Although DNNs are effective, they come at the cost of significant compute complexity, even for an image.
Evidently, passing all the frames of a video through DNN inferencing is computationally prohibitive.
To address this challenge, recent works~\cite{noscope, tahoma, blazeit, smol, focus:osdi:2018, vstore, svq, video-monitoring-queries, deluceva, panorama, vaas} have focused on \emph{cascade} architectures. They stage processing as (relatively) inexpensive predicates to filter the incoming frames of video by trading analysis accuracy for higher throughput.
As such, only a handful of frames arrive at the last stage that performs the full DNN inferencing.
\nocite{probabilistic-predicates, chameleon}

\begin{figure}[t]
	\centering
	\includegraphics[width=1.0\linewidth]{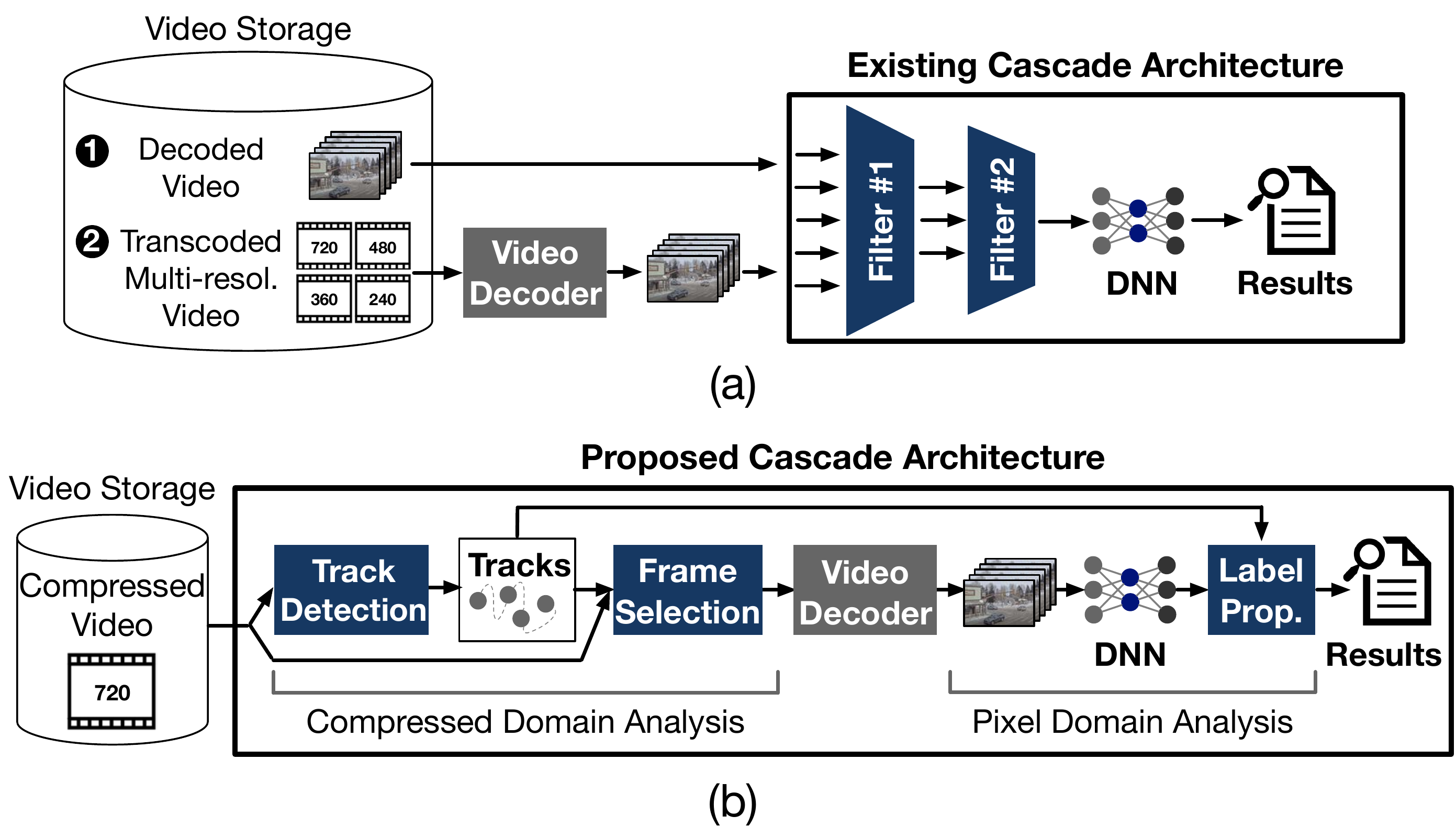}
    \caption{(a) Existing state-of-the-art cascade systems~\cite{noscope, tahoma}, excluding video decoding from the end-to-end setting with two costly assumptions; (b) the proposed cascade architecture that addresses the decoding bottleneck and supports spatial queries, exploiting the compressed domain analysis.}
	\label{fig:cascade-video-analysis}
\end{figure}

While effectively resolving the DNN throughput bottleneck, the existing cascade systems have two limitations.
First, as shown in Figure~\ref{fig:cascade-video-analysis}(a), these systems either ignore or sidestep a new bottleneck stage, \emph{video decoding}, by making one of the two costly assumptions: (1) input video is decoded a priori and the raw frames are stored in storage~\cite{noscope, tahoma, probabilistic-predicates, video-monitoring-queries}, or (2) input video is pre-transcoded and stored in multiple lower resolutions at ingest time to facilitate the query time decoding~\cite{smol, vstore}.
However, in practice, decoding (or transcoding) the entire video corpus and storing the uncompressed (or duplicate) data in disk is often infeasible due to the significant compute and storage cost.

Second, to achieve otherwise-unachievable throughput, the existing cascade systems often exclusively support \emph{temporal} queries.
More specifically, many cascade systems~\cite{noscope,tahoma,probabilistic-predicates,deluceva} only support binary predicate query, which is to get timestamps of frames that contain the queried object.
However, recent studies in video analytics~\cite{video-monitoring-queries, tasm} propose \emph{spatial} queries (e.g., car in upper right region) and demonstrate their usefulness, which cannot be supported by the current cascades.

To tackle the two limitations, this paper sets out to devise \cova\footnote{\textbf{\textsf{CoVA}}: \textbf{Co}mpressed \textbf{V}ideo \textbf{A}nalytics.}, an alternate cascade architecture.
As illustrated in Figure~\ref{fig:cascade-video-analysis}(b), the key contribution of \cova is to split cascade computation between compressed domain and uncompressed pixel domain, which collaboratively alleviate the decoding bottleneck at query time without requiring any pre-processing and support both temporal and spatial queries.
To design this cascade architecture, we leverage the following two insights:
\begin{description}[labelindent=0.5em,leftmargin=2.0em]
\item[(1)]
A small set of encoding metadata, commonly used by modern video codecs, provides noisy, yet rich, information to accurately locate potential objects and track them across frames in compressed video, while decoded pixel data is only necessary to classify objects.
\item[(2)] Video analytics queries can be fulfilled by answering the following three questions: (1) where and when are interesting objects present in the video (i.e., spatiotemporal information); (2) what are the object classes (i.e., label information); and (3) what specific information do queries ask about these objects?
\end{description}

With these insights, \cova divides video analytics over compressed footage into three major stages.
The first stage (\textbf{Track Detection}) detects occurrences of moving objects (called \emph{blobs}) over a collection of consecutive compressed frames (called \emph{tracks}).
To realize this objective, we devise a novel compressed-domain blob tracking technique, refitting a neural network based segmentation algorithm and a multiple object tracking algorithm, both of which are originally designed for pixel domain.
Our second stage (\textbf{Frame Selection}) avoids decoding the whole track and selects a minimal set of frames that are representative and yet minimize the decoding load.
\cova passes only this subset through the full DNN object detection.
The third stage (\textbf{Label Propagation}) takes the labels and the coordinates of the detected objects in the subset and uses spatiotemporal information from the first stage to propagate labels across all the frames of the track.
Altogether, these approaches offer a novel cascade architecture that performs its first and second stages in the compressed domain, while the third stage is in the pixel domain.

Finally, the three stages produce a collection of analysis results for each frame, which include a list of present objects, their pixel coordinates, their labels (e.g., car), and all other properties associated with the objects (e.g., color).
Note that the results are query-agnostic and not specific to a certain query.
Therefore, \cova runs the three stages only for the initial query and stores the analysis results along with the video in database.
When other queries are requested over the same video in a future, \cova simply retrieves the results and process the queries without reprocessing the video.

We prototype a \cova system\footnote{Our prototype is available at https://github.com/casys-kaist/CoVA.} on NVIDIA's streaming analytics framework, DeepStream~\cite{nvidia-deepstream}.
We evaluate the effectiveness of \cova using five video streams and four queries.
Compared to existing cascade systems for query time retrospective analytics, \cova offers \bench{4.8$\times$} throughput improvement, while compromising only modest accuracy loss.
We also show that \cova is capable of serving spatial queries without having significant accuracy loss, compared to the full DNN analytics baseline.

\vspace{2ex}
\noindent\textbf{Contributions.} Our key contributions are as follows:
\begin{itemize}[labelindent=0.5em,leftmargin=2.0em]
\item We show that encoding metadata is sufficiently rich to identify objects of interest along with their spatiotemporal information for retrospective video analytics.
\item To extract the spatiotemporal information, we devise a novel compressed-domain blob tracking technique, refitting the pixel-domain video segmentation and object tracking algorithms.
\item We present the design of \cova, a mixed-domain retrospective analytics system that leverages the track information to alleviate the decoding bottleneck, and support both temporal and spatial queries.
\item Our experiment shows that \cova offers significant throughput improvement over conventional cascade systems, while compromising modest accuracy loss.

\end{itemize}

 \section{Background and Motivation}
\label{sec:back-motiv}

\cova aims to tackle limitations of existing retrospective analytics systems.
Below, we first provide background on state-of-the-art retrospective analytics and discuss their limitations.
We also discuss common compression mechanisms of modern video codecs, which drive the design of proposed techniques.

\subsection{Retrospective Analytics}
\label{sec:dnn-video-analytics}

Modern retrospective analytics systems~\cite{noscope, tahoma, probabilistic-predicates, deluceva, vstore, chameleon, svq, panorama, video-monitoring-queries, vaas, smol, blazeit} share two common properties: (1) heavy reliance on DNNs, and (2) cascade architecture to resolve the DNN compute bottleneck.
While they have these common properties, there are two different dimensions that categorize the instances of retrospective analytics systems.

\niparagraph{Time of analysis -- query time vs. ingest time.}
Retrospective analytics systems are categorized into two groups, depending on whether the analysis occurs at \emph{query time}~\cite{noscope, tahoma, deluceva} or \emph{ingest time}~\cite{vstore, smol, blazeit, vaas}.
While ingest time analysis leverages offline pre-processing to facilitate and expedite the query time analysis, it requires to scan the entire video data corpus and consume compute resources on it, even though a significant portion of the data is not queried.
This approach is not only cost-inefficient but also environmentally suboptimal since it would consume a massive amount of energy for mostly unnecessary computations.
In contrast, query time analysis performs the full analysis at query time without having any pre-processing.
Therefore, it does not touch raw video data unless it is queried, which allows analysts to prevent the waste of resources.
To this end, this work focuses on the query time analysis and aims to address its limitations.

\niparagraph{Supported query -- temporal vs. both temporal and spatial.}
Most, if not all, of query time cascade systems~\cite{noscope, tahoma, deluceva} limit the types of supported queries to be only the \emph{temporal} ones and specialize the cascade stages for a specific temporal query to achieve high throughput.
However, recent work~\cite{video-monitoring-queries} points out that \emph{spatial} information can enable richer capabilities for video analytics.
\cova is a novel cascade architecture that leverages compressed-domain analysis to address both spatial and temporal queries.

 \begin{figure}
	\centering
	\includegraphics[width=0.9\linewidth]{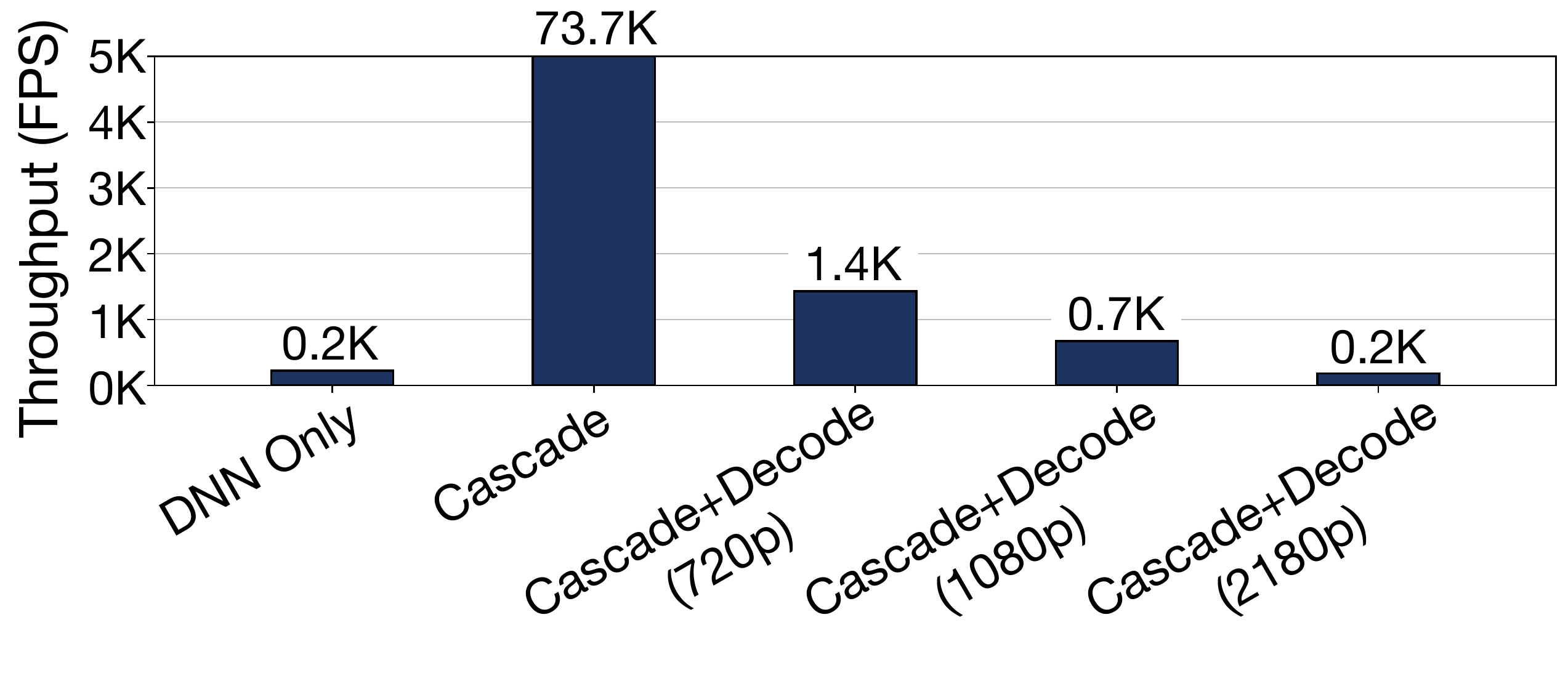}
	\caption{Throughput comparison among various system environments of cascade video analytics.\vspace{2ex}}
	\label{fig:motivation-results}
\end{figure}

\subsection{Video Decoding: the New Bottleneck}
\label{sec:decode-bottleneck}

\niparagraph{Decoding for end-to-end cascade.}
With the volume of video data growing at an explosive rate, the use of compression is imperative to keep storage costs in check.
Video codecs such as H.264 strike a balance between quality and storage size, being used as the de facto way of storing large corpus of video data.
As such, the first step in an end-to-end system for processing video queries is to decode the video data before further processing.
However, even with hardware-acceleration for standard codecs baked-in to modern CPUs and GPUs, video decoding can be up to orders-of-magnitude slower than the capabilities of cascade systems to process raw video frames.

\niparagraph{Bottleneck analysis.}
To quantify this bottleneck, we examine the performance impact of video decoding for an existing state-of-the-art cascade system, Tahoma~\cite{tahoma}, using NVIDIA RTX 3090 GPU, and present the results in Figure~\ref{fig:motivation-results}.
The detailed methodology is provided in Section~\ref{sec:method}.
The cascade system is effective in addressing the DNN-execution bottleneck and offers up to \bench{327$\times$} improvement in performance compared to a native DNN-only solution.
However, even with decoding accelerator hardware NVDEC~\cite{nvidia-nvdec}, the decoding throughput is significantly lower than the throughput of cascade system, which curtails most performance gains.

Further, as video resolution increases, the decoding throughput almost linearly decreases, exacerbating the decoding bottleneck.
Considering the trend that even IoT devices such as surveillance cameras produce HD (1080p) or higher resolution video, we believe that this decoding bottleneck will become increasingly severe and significantly hinder the usefulness of video analytics in interactive applications.
Motivated by these insights, the objective of \cova cascade is to address the decoding bottleneck in query time retrospective analytics.

\subsection{Block-based Video Coding}
\label{sec:compression-algorithms}

To alleviate the decoding bottleneck, \xvdec leverages the unique characteristics of \emph{block-based} compression, used in many modern video codecs.
Below, we provide background on block-based compression and discuss opportunities that it opens for compressed-domain analysis.

\niparagraph{Video codecs.}
Many video codecs, such as H.264, HEVC, VP8, VP9, and AV1, use block-based compression algorithm.
In this paper, we primarily focus on the H.264 format since it is one of the most widely used codecs in various applications as of publication date~\cite{wiegand:2003:overview-of-h264}.
However, \cova is compatible with other block-based codecs since all of them compress video, generating the same set of metadata we use for compressed-domain analysis in \cova.

\begin{figure*}[t]
	\centering
	\includegraphics[width=1.0\linewidth]{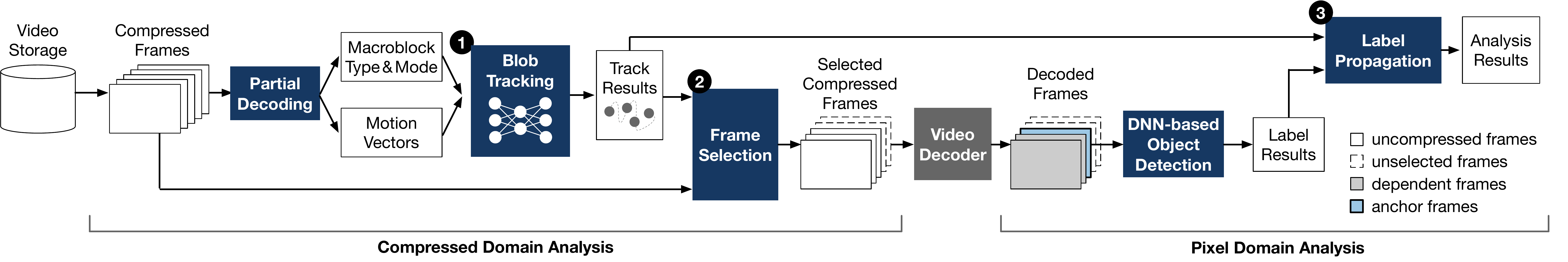}
	\caption{Overview of \xvdec.\vspace{4ex}}
	\label{fig:overview}
\end{figure*}

\niparagraph{Block-based compression.}
Block-based codecs compress (or encode) video frames by splitting each frame into a two-dimensional array of fixed sized blocks, called \textit{macroblocks} (e.g., 16x16 pixels).
There are three \underline{\textit{macroblock types}} -- I, P, and B -- depending on the way how the macroblocks are compressed.
An I-macroblock is independently compressed, while P- and B-macroblocks are compressed referring to one and two other macroblocks, respectively.
To maximize compression ratio, the codecs select dependent macroblocks for P and B-macroblocks with the highest similarity and store the spatial offsets as metadata called \underline{\emph{motion vectors}}.
Depending on the composition of macroblocks, frames are again categorized into three types, I, P, and B.
An I-frame, also known as a keyframe, is only composed of I-macroblocks, while a P-frame contains I/P-macroblocks and a B-frame has all of the I/P/B-macroblocks.

To maximize the compression rate, codecs can \textit{partition} macroblocks (e.g., 16x16) into smaller sub-macroblocks (e.g., 4x4).
This optimization allows codecs to achieve a higher compression rate but at the expense of storing larger metadata.
Modern codecs employ multiple \underline{\textit{macroblock partitioning modes}}. For instance, H.264 uses six modes from no partitioning (i.e., 1 macroblock of size (16$\times$16)) to 16-way partitioning (i.e., 16 sub-macroblocks of size (4$\times$4)).

\finding{\cova leverages the insight that the encoding metadata -- (1) macroblock types, (2) motion vectors, and (3) macroblock partitioning modes -- in the compressed video is sufficiently rich to detect potential objects and track them across frames.
}

\niparagraph{Compression rate optimization.}
Due to the higher compressibility, codecs tend to prefer P/B-macroblocks over I-macroblock.
However, the preference for P/B macroblocks ends up creating long dependency chains among the macroblocks, which cause compression errors to propagate across the chains and hinder random access to frames in the video.
To resolve the problems, the codecs insert I-frames at regular intervals, typically every 250 frames, to create independent sets of consecutive video frames, called Groups of Pictures (GoP).
Within a GoP, the number of dependent frames that need to be decoded grows linearly, with zero for the first I-frame and maximum for the last frame.

\finding{\cova exploits the inter-frame dependencies and object track information extracted from compressed-domain analysis to prudently select the frames with the least number of dependencies in each GoP that enable to identify all the objects present and minimize decoding effort.
}

 \section{Overview of \xvdec}
\label{sec:overview}
\cova divides video analytics over compressed footage into three major stages, as illustrated in Figure~\ref{fig:overview}.

\circledparagraph{1}{First Stage: Track Detection.}
First, \cova detects occurrences of moving objects over a collection of consecutive compressed frames, which we call tracks.
The track detection stage further breaks down into two steps: (1) \emph{blob detection}: \cova \emph{spatially} detects whether and where moving objects (called blobs) are present in each compressed frame; and (2) \emph{blob tracking}: \cova \emph{temporally} associates the blobs across frames to identify unique blob tracks.
For the blob detection, we devise a novel compressed-domain blob detection model, refitting a neural network architecture originally designed for pixel-domain video segmentation.
The neural network only takes as input three encoding metadata commonly used by modern codecs, recognizes movements as masks, and spatially associates the masks clustered in a region as blobs.
While the neural network architecture is fixed, \cova trains the model individually for each video to learn the data-specific patterns of blobs and specialize for the target video.
Finally, the found blobs are fed into the blob tracking step that employs an object tracking algorithm, SORT~\cite{bewley:2016:sort}, which was also originally developed for pixel domain.
Note that the blob track results still lack the object class labels.

\circledparagraph{2}{ Second Stage: Frame Selection.}
To attain the object labels for the detected blobs, \cova needs to perform DNN-based object detection for the frames where tracks appear, which ordinarily require decoding all the frames.
However, as frames on a track most likely contain the same object, it is enough to perform the object detection on a subset of the frames in the track, which we call  \emph{anchor frames}.
Thus, \cova only decodes frames required to decode the anchor frames, which improves the effective decoding throughput.
The challenge is how to prudently select the anchor frames so as to minimize the decoding cost and at the same time acquire the accurate label information.
We develop a frame selection algorithm that leverages a common property of video codecs where compressed frames are encoded in dependency chains.
Thus, anchor frames are the ones that are located on the maximal number of tracks and at the same time have the short dependency chain with respect to the decoding algorithm.
Note that while the anchor frames are the only ones that are inferred upon for object detection, all the frames in the track need to be labeled to handle various video analytics queries.

\circledparagraph{3}{Third Stage: Label Propagation.}
In the third stage, \cova takes the approximate positions of potential objects (or blobs) from the first stage and labels for the anchor frames from the second stage to temporally propagate the labels across all the frames of the tracks.
To merge the spatial and temporal results, \cova first spatially correlates blobs with objects on anchor frames using the intersection ratios of their bounding boxes.
Then, \cova uses the tracking information to identify the same objects across the frames and propagates the labels, while populating bounding boxes around the corresponding blobs in the temporally consecutive frames.

Finally, when a video passes through the three stages, \cova produces a collection of analysis results for each frame, the examples of which are a list of present objects, their pixel coordinates, their labels (e.g., car), and all other properties associated with the objects (e.g., color).
Note that the results are created only once when \cova receives the initial query over a video and they are permanently associated with the video in the database.
After then, analysts can use the same results to process various future queries without reprocessing the video.

\section{Compressed Domain Blob Tracking}
\label{sec:cdbt}

\begin{figure}
    	\centering
	\includegraphics[width=0.9\linewidth]{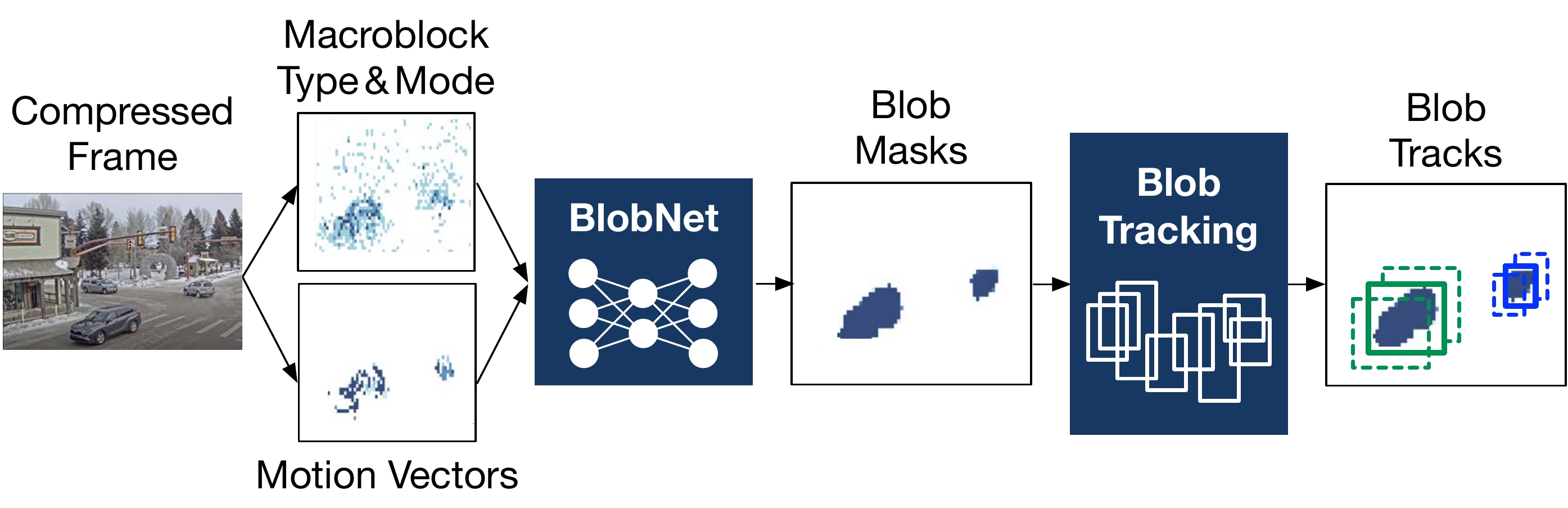}
	\caption{Track detection.\vspace{2ex}}
	\label{fig:blob-tracking}
\end{figure}

In this section, we describe the track detection mechanism that is the first stage of \cova's cascade architecture.
Figure~\ref{fig:blob-tracking} depicts the overall workflow.

\subsection{Learning to Detect Blobs}

\niparagraph{Limitations of existing compressed domain video processing techniques.}
Detecting objects or blobs from compressed video is a traditional research problem in the computer vision community~\cite{alizadeh:2019, zeng:2005, babu:2004:cd-vos, laumer:2016, poppe:2009, nguyen2020toward}.
However, the following two limitations prevent the simple adoption of these techniques.
First, the techniques often require human-crafted parameters that need to be tuned for each input video, which makes automated analytics impossible.
Secondly, the techniques are not sufficiently robust to be applied to arbitrary video data, producing inadequately accurate tracking results for video analytics.
To overcome such limitations, recent works~\cite{coviar, mmnet} explored to use neural networks for vision tasks over compressed video.
Unfortunately, we could not employ the neural networks for \cova since they not only still require pixel-domain data for a subset of frames, but also offer insufficient throughput that is significantly lower than the decoder.

\niparagraph{Leveraging the similarity between video segmentation and blob detection.}
To address these limitations, we exploit an observation that blob detection using compression metadata is akin to the problem of the semantic image segmentation using pixel data.
\emph{Blob detection} task aims to find potential objects and their approximate position within video frames.
\emph{Image (or video) segmentation} task, on the other hand, aims to semantically split an image (or frames of a video) and classify each segment into one of the predetermined labels.
When there are only two classes -- blob and non-blob -- the image segmentation task can be reduced to the approximate blob detection task.
This observation allows us to tap into the vast range of techniques, including Deep Neural Network (DNN) based image and video segmentation, that can be geared towards compressed domain blob detection.

\subsection{BlobNet}
To this end, we devise a lightweight DNN-based blob detection model, called \blobnet, building upon the state-of-the-art Temp-UNet~\cite{tempunet} model for video segmentation.
Unlike the Temp-UNet model, which operates on pixel frames, \blobnet operates on compression metadata.

\begin{figure}
    	\centering
	\includegraphics[width=1.0\linewidth]{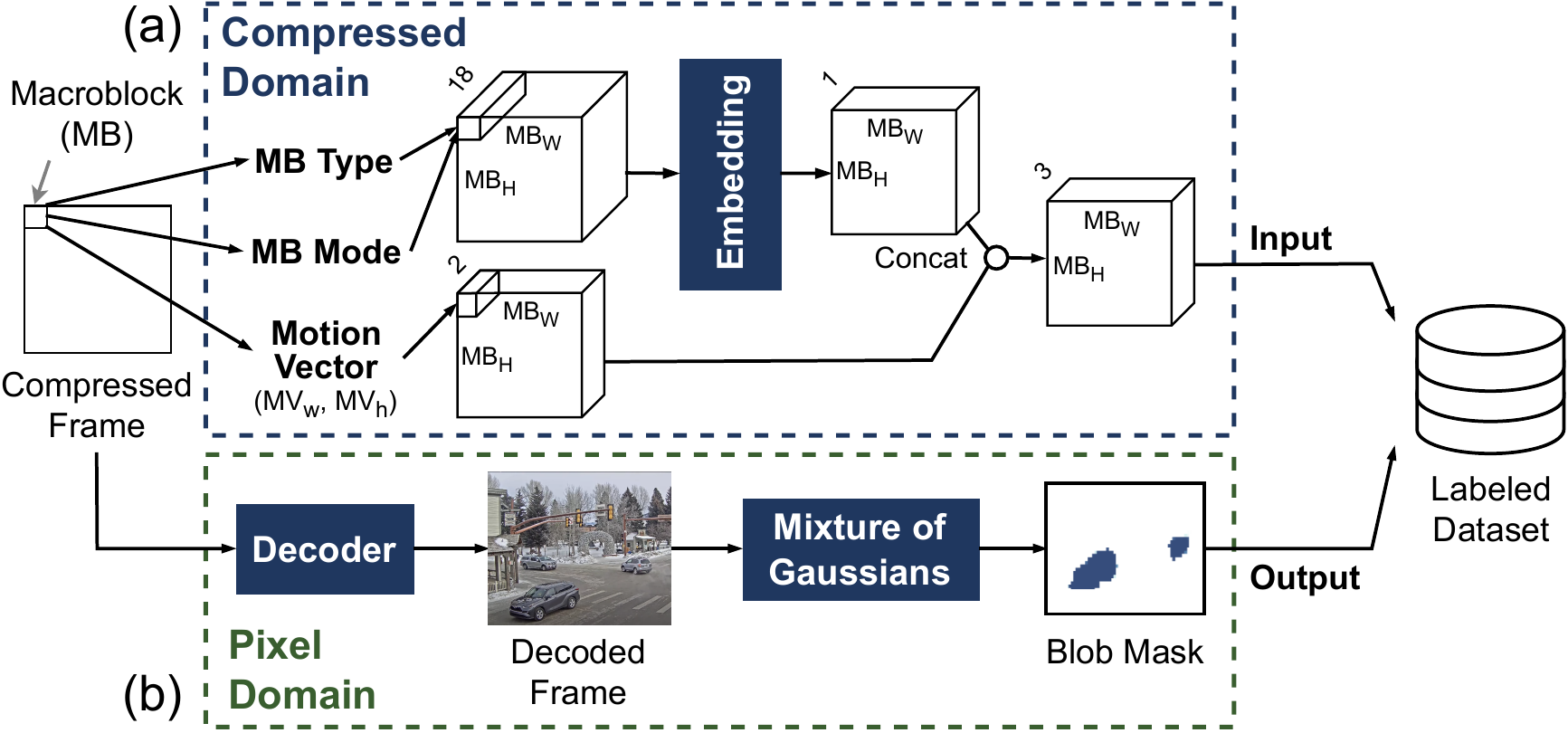}
	\caption{(a) Feature engineering that transforms three compression metadata into a tensor of input features; (b) labeled data collection using the Mixture of Gaussians (MoG) model.\vspace{1ex}}
	\label{fig:blobnet-training}
\end{figure}

\niparagraph{Feature engineering.}
Figure~\ref{fig:blobnet-training}(a) depicts the feature engineering, which converts the three metadata into a tensor of input features.
\blobnet takes the three types of metadata as input -- macroblock types, macroblock partitioning modes, and motion vectors.
To obtain the metadata, \cova performs only a few early stages of the decoding process required to extract metadata, called \emph{partial decoding}.
\cova encodes the first two metadata, macroblock types and partitioning modes, by mapping each of their combinations into an one-hot vector (e.g., total 12 combinations for H.264).
These one-hot vectors are fed into an embedding layer, which converts each one-hot vector into a scalar weight value.
This weight value is concatenated to the motion vector ($MV_w$, $MV_h$) for each macroblock, which finally results in a 3D tensor ($MB_{W} \times MB_{H} \times 3$).
\cova temporally stacks these tensors from consecutive frames and constructs a 4D tensor, which is the input for \blobnet.

\niparagraph{\blobnet architecture.}
Similar to the architecture of Temp-UNet\footnote{We omit the detailed architecture and refer to the paper~\cite{tempunet}.}, \blobnet has three major components: (1) \textbf{encoder} that extracts the presence and approximate location of blobs from noisy metadata; (2) \textbf{decoder} that reconstructs the shapes of blobs from the blob presences; (3) \textbf{skip connections} that offer spatial information to the decoder for assisting the shape reconstruction process.
While this overall composition is the same as that of Temp-UNet architecture, we maximally reduce the depth of encoder and decoder modules such that the resulting model still offers high accuracy while maximizing the inference throughput.

\niparagraph{Video-specialized model training.}
Pixel video segmentation models typically train once during a training phase, followed by inference on unseen video data.
However, \cova trains \blobnet at query time for every video data to specialize the model for the specific data.
This design choice is derived from our empirical observation that without such model specialization, the model cannot capture the variations of data-specific encoding parameters and fails to reach sufficient accuracy.
Note that once training is completed for a video data, no further training is required for additional video if the video is recorded from the same angle of view with the trained one.
We empirically observe that $\approx3\%$ of the video is sufficient to train the model for the evaluated video (see Table~\ref{tab:dataset}).
The training process, including data collection and training, takes only a few minutes, which can be amortized for multiple queries on the same video data.
Such training cost amortization is inspired by existing query-time cascade systems~\cite{noscope,blazeit,smol,tahoma} that train specialized neural networks for each video.

\niparagraph{Labeled data collection for supervised learning.}
As \cova aims for large-scale video analytics, manually labeling the video data is infeasible.
As such, \cova needs a method to automatically label the video data.
Similar to prior works~\cite{noscope,blazeit,tahoma, adaptive-cascading:cvpr:2017}, using pixel domain object detection is a possible option.
However, object detection models are not only computationally expensive but also produces labels for non-moving objects, which should not be used to train \blobnet, designed to detect only moving objects.
Instead, we exploit the conventional Mixture of Gaussians (MoG) based background subtraction technique since it is lightweight and only looks for the moving objects.

\subsection{Tracking Blobs}
\label{subsection:blob-tracking}

\niparagraph{Blob detection results.}
The output of \blobnet is merely a collection of 1's on the resulting bitmap, which lacks the notion of objects.
\cova uses connected-component labeling algorithm to uniquely identify the interesting regions in compressed frames as potential objects, called \textit{blobs}.
Once the blob identification process is completed, \cova obtains the spatial information of blobs on each frame.
However, the blobs existing across consecutive frames are not yet temporally associated with each other, which necessitates the next stage of \cova, blob tracking.

\niparagraph{SORT-based blob tracking.}
The end objective of blob tracking in compressed domain is to minimize the number of frames to be decoded to mitigate the decoding bottleneck.
Hence, the tracking algorithm must (1) offer high throughput that significantly outperforms the decoder throughput, (2) while accurately tracking the inter-frame blobs to minimize the accuracy loss at the label propagation stage.
We extensively explore existing object tracking techniques in pixel domain~\cite{bae:2014:robust-mot-tracklet, yang:2016:temporal, xiang:2015:learning-to-track, bewley:2014:online-multi-instance-segmentation, choi:2015:near-online-multi-target-tracking, yoon:2015:bayesian-multi-object-tracking, bewley:2016:alextrac, bewley:2016:sort}, and choose the SORT object tracking algorithm~\cite{bewley:2016:sort}, which satisfies the above two requirements.
SORT offers the near-best tracking accuracy among the state-of-the-art tracking techniques and
keeps the computation lightweight by exploiting conventional optimization algorithms, Kalman filter and Hungarian assignment.

 \section{Track-aware Frame Selection}
\label{sec:filtering}

\begin{figure}
    	\centering
	\includegraphics[width=0.95\linewidth]{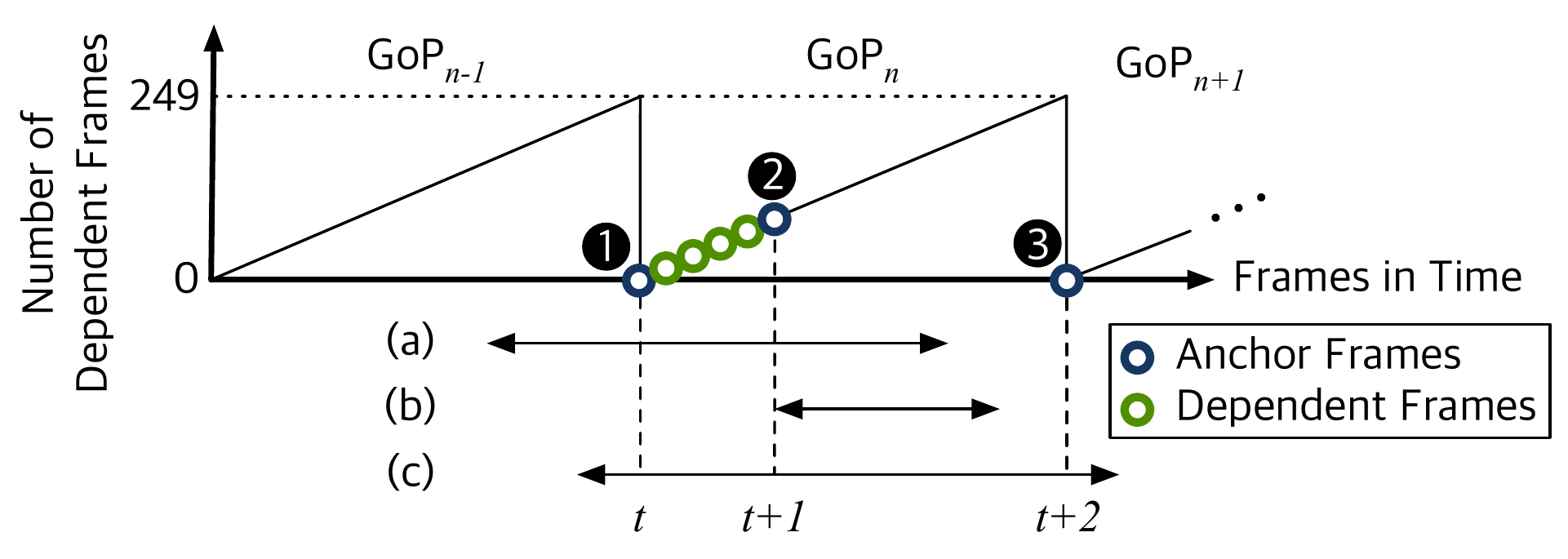}
	\caption{Example scenario of track-aware frame selection.}
	\label{fig:frame-filtering}
\end{figure}

Leveraging the track information, \cova prudently select a small subset of frames to decode, called \textit{anchor frames}, so as to maximize the decoding throughput.
The key idea behind the anchor frame selection algorithm is to pick the ones that require to decode the least number of frames and thus maximize the \textit{effective} decoding throughput.

\niparagraph{Dependency between compressed frames.}
As described in Section~\ref{sec:compression-algorithms}, block-based compression uses a combination of (1) independent frames that are self-contained (i.e., I-frame), and (2) dependent frames (i.e., P/B-frames) that depend on either preceding frames, subsequent frames, or both.
Due to the presence of P-frame and B-frame within a GoP, the number of dependent frames that need to be decoded to fully decode a frame follows a saw-tooth structure, as depicted in Figure~\ref{fig:frame-filtering}.
The number of dependent frames is zero for I-frame at a GoP boundary and grows linearly until it resets to zero at the end of GoP\footnote{For brevity, we simplify Figure~\ref{fig:frame-filtering} by only visualizing dependency chains for P-frames since the number of dependent frames for B-frames is similar to that of the nearby P-frames.}. 

\begin{algorithm}[t]
\LinesNumbered
\fontsize{8.5}{10.0}\selectfont
\tline
\vspace{2pt}
    \SetKwInOut{Input}{Input}	
    \SetKwFunction{Deocde}{Decode}
	\SetKwInOut{Output}{Output}
 
    \Input {efs: compressed frames in a GoP$_t$ \\
    		tracks: blob tracks that maintain across GoPs
    }
    \Output {dfs: compressed frames chosen to be decoded \\
    		 afs: anchor frames}
    
\vspace{1ex}
\normalem

cur\_tracks = tracks that terminate in GoP$_t$ \\
\quad \quad \quad \quad \quad \quad \quad \quad with no anchor frames assigned\\
dfs = afs = $\emptyset$ \\ 
\If{\emph{cur\_tracks} $\ne \emptyset$}
{
	start\_timestamps = \textbf{sorted}(cur\_tracks.starts()) \\
	end\_timestamps = \textbf{sorted}(cur\_tracks.ends()) \\
	sidx = eidx = 0 \\

	\For{\emph{ef} \textbf{\emph{in}} \emph{efs}}{
		\While{\emph{start\_timestamps[sidx] \textbf{==} ef.timestamp}}{
			candidate\_af = ef \\
			sidx = sidx + 1 \\
		}
		\While{\emph{end\_timestamps[eidx] \textbf{==} ef.timestamp}}{
			afs.add(candidate\_af) \\
			dfs.add\_dependants(candidate\_af, efs) \\
			eidx = eidx + 1 \\
		}
	}
}
dfs.output() \\ 
afs.output() \\

\vspace{1.5ex}
\bline
\vspace{1ex}
\caption{\textbf{Track-aware frame selection algorithm.}}
\label{alg:frame-filtering}
\end{algorithm}

\niparagraph{Selecting anchor frames for decoding.}
To minimize the decoding load, we leverage two insights: (1) \cova can find the consecutive frames where an object keeps appearing in the video, and (2) the computations load to decode a frame is proportional to its number of dependent frames.
Within each GoP, \cova identifies a set of anchor frames, which can identify all objects present in the GoP and perform the least computation for decoding, by minimizing the number of dependent frames.
The selected anchor frames are the only ones that are passed to the DNN object detector to produce the label information.

\niparagraph{Example.}
Figure~\ref{fig:frame-filtering} presents an example where \cova identifies three unique objects, (a), (b), and (c), as well as the range of frames where each object stays in the video.
In this example, the best choice of anchor frame would be Frame \circled{2} since (1) all the objects are present in Frame \circled{2}, and (2) Frame \circled{2} has the least number of dependent frames among frames where all the objects are present.

\niparagraph{Algorithm.}
Algorithm~\ref{alg:frame-filtering} describes the frame selection algorithm in detail.
\emph{Line 1}: When a GoP arrives at the frame filtering, to select the anchor frames, \xvdec only considers tracks that (1) terminate in that particular GoP and (2) do not have any anchor frames yet (e.g., object (a)/(b) at time \emph{t}).
\emph{Line 9}: Then, as \xvdec visits frames in order, it first checks if a track starts appearing in the visiting frame.
\emph{Line 10}: If it does, the visiting frame is marked as ``candidate'' anchor frame (e.g., Frame~\circled{1} at \emph{t}).
Later on, if a new track starts appearing in a successive frame, the frame becomes the new candidate (e.g., Frame~\circled{2} at \emph{t+1}).
\emph{Line 14--15}: When a track ends, \xvdec adds the current candidate frame into the anchor frame list (e.g., Frame~\circled{2}) and inserts all the dependent frames into the dependent frame list (e.g., all frames between Frame~\circled{1} and Frame~\circled{2}).
The intuition behind this algorithm is that, if a track started but did not terminate, any frame in between can be an anchor frame.
However, when a track ends, an anchor frame for the track must be selected, because otherwise, we may not have any anchor frame for the terminating track.

 \section{Label Propagation}
\label{sec:collaborative}

\begin{figure}
    	\centering
	\includegraphics[width=1.0\linewidth]{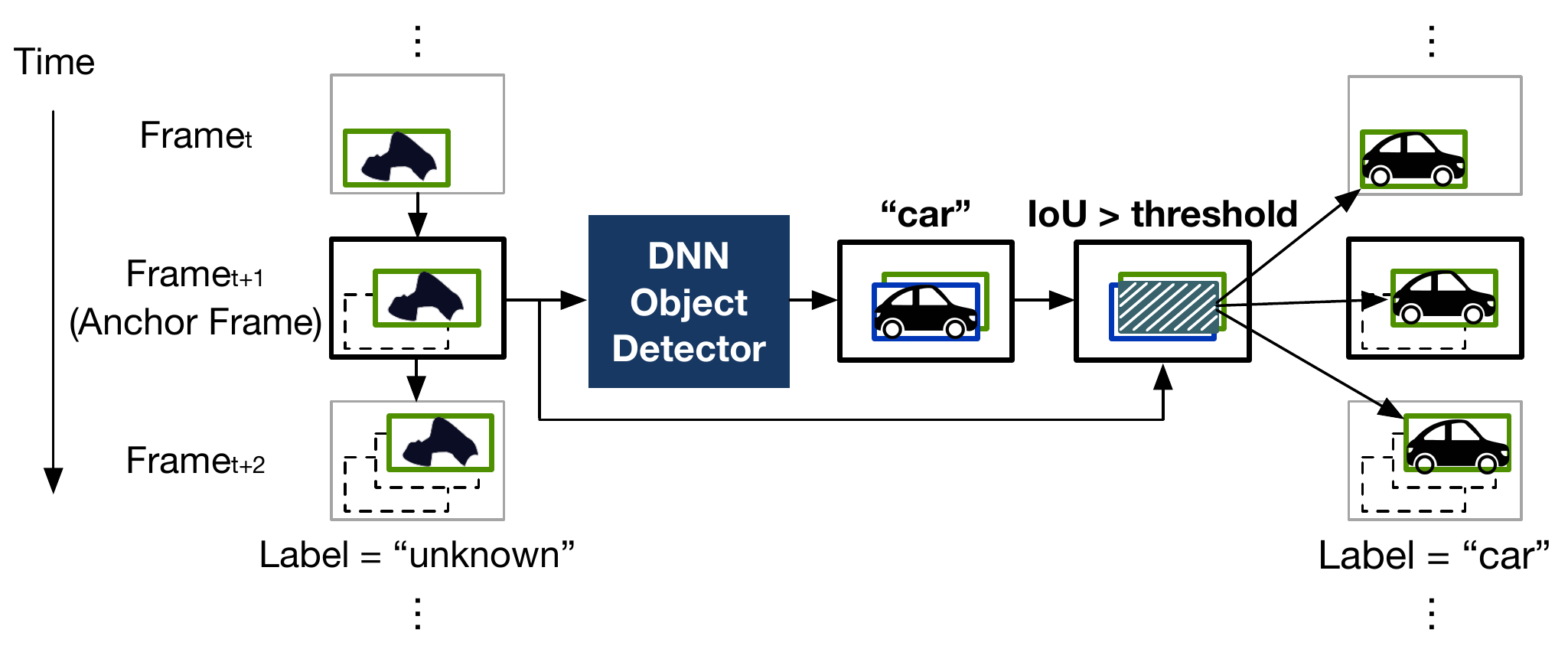}
	\caption{Label propagation.\vspace{-2ex}}
	\label{fig:collaborative_query}
\end{figure}

In the last stage, \cova takes the blob tracks and labels for the anchor frames to temporally propagate the labels across all the frames on the tracks.
Figure~\ref{fig:collaborative_query} illustrates the example workflow of label propagation.
When the selected anchor frames and their dependent frames are decoded, \cova takes only anchor frames to perform the DNN object detection and obtain the labels (e.g., ``car'') as well as their spatial information.
To associate the labels with blobs, \cova first spatially correlates blobs with the detected objects using the intersection over union (IoU) between their bounding boxes (e.g., bounding boxes of blobs and detected objects are denoted using green and blue boxes, respectively).
When the IoU is larger than a threshold, \cova associates the detected objects with blobs and propagates the labels to all frames in the tracks.

\niparagraph{Multiple-objects overlapping problem.}
One challenge with the label propagation mechanism is that when BlobNet fails to separately identify multiple objects clustered together and creates a large single blob, \cova cannot correctly propagate the multiple labels.
To overcome the challenge, we prepend an additional step to the label propagation.
When a multitude of detected objects are spatially overlapped with a single blob, \cova splits the blob into multiple blobs, proportionally projecting the locations of objects in the anchor frame to the blob.
The proportional projection is also applied to other frames in the same track, populating multiple tracks from a single track.
This way, \cova is able to propagate the multiple labels to the separated tracks, instead of giving a single erroneous label to the clustered objects.

\niparagraph{Static object handling mechanism.}
As \cova relies on the compressed domain analysis to detect blobs, it is impossible to detect static objects from the compression metadata.
Therefore, our \blobnet focuses on detecting moving objects, intentionally excluding the static object information from the training data through the use of MoG.
However, \cova still performs full-fledged object detections on anchor frames.
Therefore, the static objects can be detected at least on the anchor frames.
As the static objects stay still at the same location across subsequent anchor frames, \cova is able to associate them as the same object and produce the corresponding track.

  \section{Implementation}
\label{sec:impl}

\niparagraph{System architecture and constituent software modules.}
We prototype a \cova system using DeepStream, which is built upon GStreamer, for constructing the skeleton pipeline of video analytics. 
As described in Section~\ref{sec:cdbt}, the initial stage of \cova is the partial decoding, which extracts the metadata. 
Hardware-accelerated decoder (e.g., NVDEC) does not support partial decoding and only generates the fully decoded frames. 
Thus, we modify an open-source video codec, libavcodec, such that it only produces the three types of metadata. 
In addition, \cova performs two neural network inferences, one for the blob detection and the other for the full DNN inference (YOLOv4).
We use on a TensorRT-based DNN inference plugin on DeepStream, nvinfer~\cite{nvinfer}.

\niparagraph{Parallelization in \cova.}
Our prototype system distributes the computations of pipeline stages over CPU and GPU, while exploiting their parallelism. 
Initially, \cova scans the entire video and splits it into chunks at the I-frame boundaries to parallelize the computation on CPU threads. 
This scanning takes just a few seconds even for hours of video data, which imposes negligible overhead. 
Such parallelization results in cutting tracks at the chunk boundaries, but its impact on accuracy is negligible since there are only a few dozens of chunks. 
For a chunk, the first two stages, track detection and frame selection, should be pipelined in the same thread since these algorithms rely on the temporal dependencies of frames.
For object detection, anchor frames are independently computed, which can be fully parallelized. 
Therefore, \cova maintains only a single thread for object detection and anchor frames from different chunks are batched together for inference.

 \section{Evaluation}
\label{sec:result}

\subsection{Methodology}
\label{sec:method}

\niparagraph{Queries.}
To demonstrate the effectiveness of \xvdec, we evaluate four example queries, two queries widely used in prior work~\cite{noscope, blazeit, tahoma}, and their spatial variants supported by \cova.
Table~\ref{tab:query} reports the list of evaluated queries with their descriptions and accuracy metrics:

\begin{table}[t]
       \small
       \caption{Descriptions of example video analytics queries.}
       \label{tab:query}
       \centering
       \begin{tabular}{C{1.2cm}C{1.0cm}L{3.2cm}C{1.4cm}}
				\hlinewd{0.7pt}
				\textbf{Query} & \textbf{Abbr.} & \centering \textbf{Description} & \textbf{Metric} \\
				\hlinewd{0.7pt}
				Binary Predicate & BP & Return frames where querying object appears & Accuracy \\
				\hline
				Count & CNT & Return the average count of querying object in frames & Absolute Error \\
				\hline
				Local Binary Predicate & LBP & Return frames where querying object appears in a certain region of frames & Accuracy \\
				\hline
				Local Count & LCNT & Return the average count of querying object in a certain region of frames & Absolute Error \\
\hlinewd{0.7pt}
       \end{tabular}
\end{table}

\begin{table*}[t]
       \small
       \caption{Descriptions of video datasets, queried objects, ground truth results, and region of interest used for spatial queries. Note that we use the Yolov4 DNN model applied frame-by-frame to the original video to get ground truth.}
       \label{tab:dataset}
       \centering
       \begin{tabular}{C{2.0cm}C{1.3cm}C{1cm}C{1.3cm}C{1.6cm}C{1.2cm}C{2.0cm}C{1.2cm}C{2.0cm}}
				\hlinewd{0.7pt}
				\textbf{Video Name} & \textbf{Num of Frames} & \textbf{Length} & \textbf{Object in Interest} & \textbf{Object Occupancy} & \textbf{Object Count} & \textbf{Local Occupancy} & \textbf{Local Count} & \textbf{Region of Interest} \\
				\hlinewd{0.7pt}
				amsterdam~\cite{dataset-amsterdam} & 3,580K & 33H & Car & 70.07\% & 1.40 & 29.05\% & 0.44 & Lower Right \\
				\hline
				archie~\cite{blazeit} & 3,567K & 33H & Bus & 10.48\% & 0.17 & 6.63\% & 0.11 & Upper Left \\
				\hline
				jackson~\cite{dataset-jackson} & 2,921K & 27H & Car & 31.91\% & 0.56 & 18.28\% & 0.29 & Lower Left \\
				\hline
				shinjuku~\cite{dataset-shinjuku} & 1,782K & 16H & Car & 82.29\% & 2.19 & 19.91\% & 0.38 & Lower Left\\
               	\hline
				taipei~\cite{dataset-taipei} & 3,564K & 33H & Car & 84.48\% & 5.03 & 22.16\% & 0.64 & Lower Right\\
				\hlinewd{0.7pt}
				\vspace{2ex}
       \end{tabular}
\end{table*} 
\begin{description}[labelindent=0.5em,leftmargin=2.0em]
\item[(1)] \textbf{Binary Predicate.}
Binary predicate (BP) query finds frames where queried objects appear.
Collecting frames with queried objects is an initial step of advanced analysis, which makes BP an important query for evaluation despite the simplicity.
Many retrospective analytics systems evaluate their solutions only using the BP query~\cite{noscope, tahoma}.

\item[(2)] \textbf{Count.}
The count (CNT) query is introduced by a prior work, BlazeIt~\cite{blazeit}, which is an aggregate query that counts the number of queried objects appearing in the whole video.
As the aggregated count is largely dependent on the length of each dataset, the number is normalized by dividing it by the number of frame counts.

\item[(3)] \textbf{Local Binary Predicate and Local Count.}
The local binary predicate (LBP) and local count (LCNT) queries are spatial variants of BP and CNT queries, respectively; however, the only difference is that they exclusively look for objects located in a certain region of interest.
For instance, users can query northbound traffic in highway monitoring video by annotating the corresponding region of video as ``northbound''.
Serving these queries not only requires the temporal query results, but also needs spatial information to determine the object locations.

\end{description}

\niparagraph{Metrics.}
Table~\ref{tab:query} also reports metrics used for each query.
We use the same metrics that prior works use to evaluate their solutions.
For BP and LBP, as in prior works~\cite{noscope, tahoma}, we use \textit{accuracy}, which is a traditional metric for binary classification that evaluates how many observations, both positive and negative, are correctly classified.
Similarly, for CNT and LCNT, we use \textit{absolute error} as used in BlazeIt~\cite{blazeit}.

\niparagraph{Datasets.}
Table~\ref{tab:dataset} reports the video datasets used for the evaluation.
Taking a similar approach with prior works~\cite{blazeit, noscope, li:2020:reducto, smol, tahoma, focus:osdi:2018, video-monitoring-queries}, we collect the video datasets from YouTube live streams~\cite{dataset-amsterdam, dataset-jackson, dataset-shinjuku, dataset-taipei}.
They are recorded from statically installed cameras, which is a widely used setup in various applications domains such as traffic monitoring~\cite{traffic-monitoring-1,traffic-monitoring-2,traffic-monitoring-3}, security~\cite{video-security, video-security2}, surveillance~\cite{evolution-video-surveillance}, and healthcare~\cite{healthcare}.
Video contents involve various kinds of scenarios, which include traffic circle, highway, harbor, city streets, and park.
As the datasets have different resolutions ranged from 720p to 2160p, we transcode them to 720p and evaluate the throughput and accuracy for ease of comparison.
Note that higher resolutions (e.g., 2160p) create more severe decoding bottleneck, so using them would be favorable to \cova, producing higher throughput gains and therefore, to be conservative, we choose to use 720p for all video datasets.
The rightmost five columns report the ground truth results for the four queries and the region of interest that spatial queries focus on.
Getting the ground truth results by manually labeling the hours of video data is infeasible, so we apply a full DNN model (YOLOv4) to the entirety of video in a frame-by-frame manner.

\niparagraph{Hardware specifications.}
Our \cova prototype is built on a server with two 16-core Intel Xeon Gold 6226R CPU (2.9 GHz), 192 GB of DRAM, and an NVIDIA RTX 3090 GPU (24 GB GDDR6 DRAM).
We turn off hyperthreading to avoid interference among threads.

\niparagraph{Decoder.}
For all experiments, we use NVIDIA's hardware accelerated decoder, \textsf{NVDEC}, for both baseline and \cova systems to make a fair comparison.
We choose not to use the CPU decoder, \textsf{libavcodec}, since it shows lower throughput than \textsf{NVDEC} even with 32-core parallelization.

\begin{figure}[t]
    	\centering
	\includegraphics[width=0.95\linewidth]{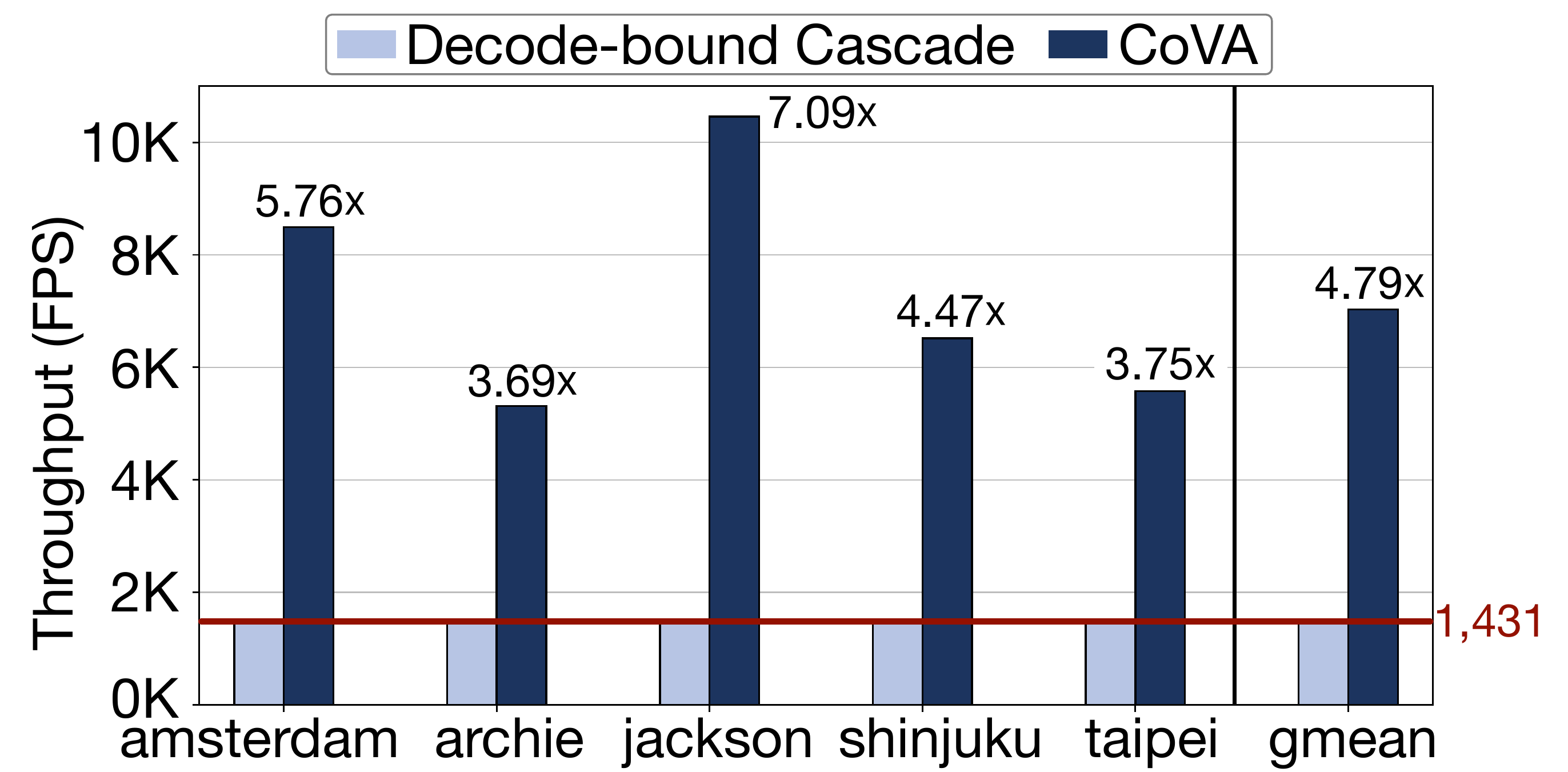}
	\caption{End-to-end system throughput of the baseline decode-bound cascade and \cova. The throughput of decode-bound cascade is equivalent to the throughput of \textsf{NVDEC} (i.e., 1,431 FPS), which is marked with a red line.}
	\label{fig:throughput}
\end{figure}

\niparagraph{Baseline cascade system.}
As the baseline, we use existing cascade systems for query time retrospective analytics. As discussed in Section~\ref{sec:back-motiv}, cascade systems such as Tahoma~\cite{tahoma} are significantly bottlenecked by video decoding.
Therefore, for a conservative comparison with these decode-bound cascade systems, we assume that the cascade systems are only bottlenecked by the decoder, not by any other stages.
With this assumption, the throughput of cascade systems is equivalent to the decoder throughput.
We refer to this baseline as \textit{decode-bound cascade} in this paper.

 \subsection{Performance Implication of \xvdec}
\label{sec:throughput-improvement}

\niparagraph{Throughput improvement.}
Figure~\ref{fig:throughput} compares the end-to-end system throughput of the baseline decode-bound cascade system and \cova across five video datasets.
\cova achieves on average \bench{4.8$\times$} throughput improvement, which ranges from \bench{3.7$\times$} for \bench{archie} to \bench{7.1$\times$} for \bench{jackson}.
The significant speedup shows that \cova effectively pushes a large proportion of analysis to the compressed domain, unclogging the decoding bottleneck that prevents the existing cascades to achieve beyond the constant \textsf{NVDEC} throughput.
The results also suggest that depending on the datasets, \cova sees different speedups.
The datasets, \bench{jackson} and \bench{amsterdam}, see relatively larger gains, while \bench{archie} and \bench{taipei} datasets show lower benefits.
These gaps can be attributed to the unique content properties of each evaluated video dataset that deliver varying throughput for the \cova pipeline stages, which eventually engenders a different bottleneck point.
To better understand the throughput implication of these stages, we delve into the interplay of algorithms and system in the \cova pipeline below.

\begin{table}[t]
       \small
       \caption{(1) Filtration rate at decoder stage (decode filtration rate) and (2) filtration rate at DNN inference stage (inference filtration rate).}
       \label{tab:filteration-rate}
       \centering
       \resizebox{0.95\linewidth}{!}{
       \begin{tabular}{C{1.5cm}C{2.4cm}C{2.7cm}}
				\hlinewd{0.7pt}
				\textbf{Dataset} & \textbf{Decode Filtration \ \ \ \ Rate (\%)} & \textbf{Inference Filtration  Rate (\%)} \\
				\hlinewd{0.7pt}
				amsterdam & 87.16 & 99.60 \\
				\hline
				archie & 72.94 & 99.15 \\
				\hline
				jackson & 94.81 & 99.79 \\
				\hline
				shinjuku & 77.18 & 99.26 \\
				\hline
				taipei & 74.03 & 99.81 \\
				\hlinewd{0.7pt}
       \end{tabular}
       }
\end{table}

\niparagraph{Effectiveness of frame selection.}
Frame selection is the key to alleviate the decoding bottleneck since it determines the computational load for decoder.
Table~\ref{tab:filteration-rate} reports the filtration rates at decoding stage (decode filtration rate) and DNN inference stage (inference filtration rate).
The decode filtration rate is calculated based on the number of decoded frames that include both anchor frames and their dependent frames, while the inference filtration rate only considers the anchor frames that are passed to the DNN object detection stage.
Intuitively, various semantics of datasets cause different filtration rates.
If video contains many objects having lots of motions, blob tracking would produce numerous tracks, which would require many anchor and dependent frames to proceed to the decoder.
For crowded video streams such as \bench{archie}, \cova sees lower decode filtration rate of \bench{72.94\%}, while the uncongested ones like \bench{jackson} capture less activity and provide higher decode filtration rate of \bench{94.81\%}.
Across all datasets, \cova filters out over \bench{73\%} to deliver over \bench{3.7$\times$} (=\bench{100/(100-73)}) throughput boost for decoder.
At the same time, the inference filtration rate closely reaches \bench{100\%}, which addresses the DNN bottleneck since the object detector only sees a handful of frames.

\begin{figure}[t]
    \centering
    \includegraphics[width=0.95\linewidth]{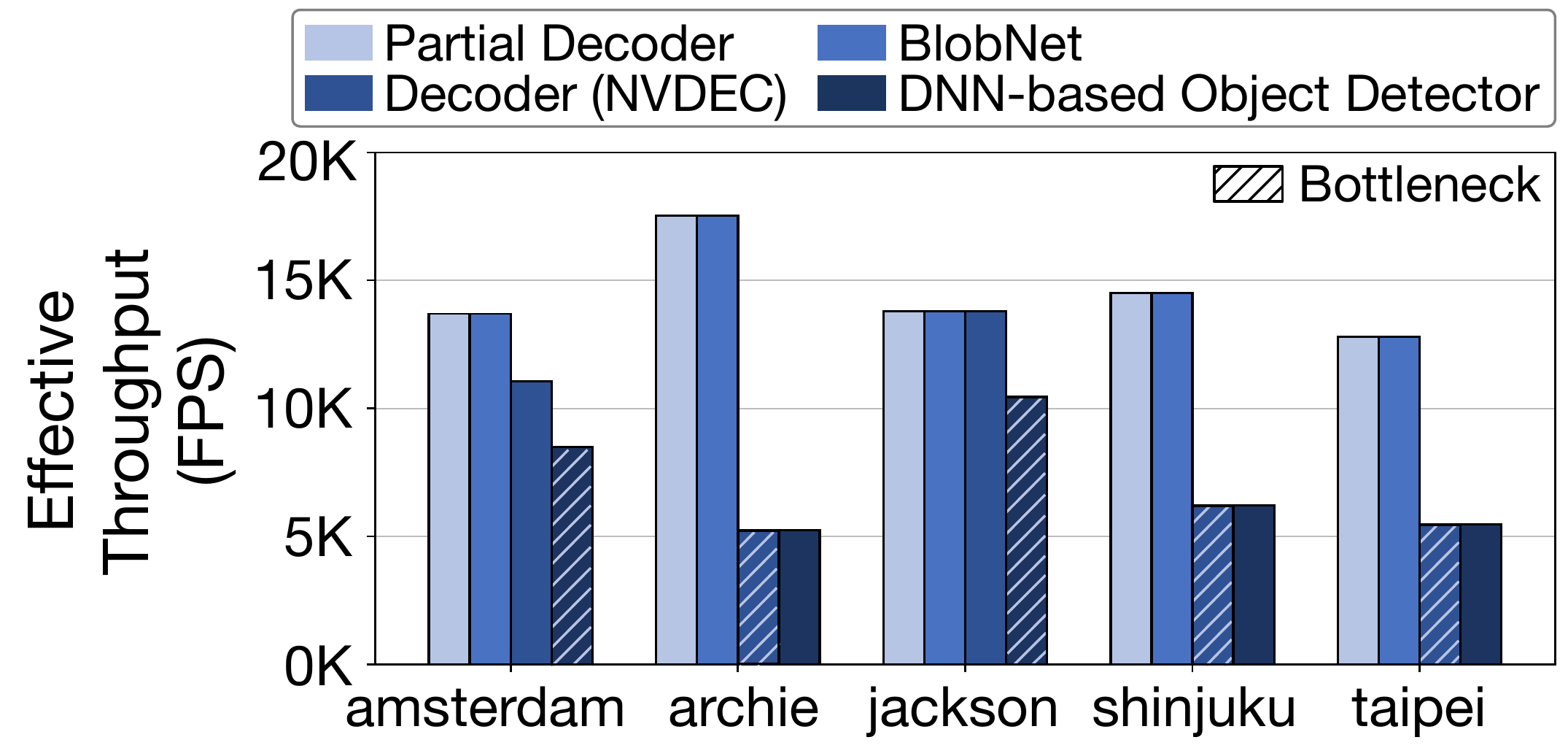}
    \caption{Effective throughput of \cova stages. The lowest bar represents the bottleneck of \cova pipeline, which is marked with hatching lines.\vspace{1ex}}
    \label{fig:throughput-cascade}
\end{figure}

\niparagraph{Bottleneck analysis.}
To understand the throughput variation of \cova stages across different datasets, we measure the performance of individual stages.
Figure~\ref{fig:throughput-cascade} reports the effective throughput of each stage by starting from the first partial decoding stage and adding successive stages one by one to the system.
The \textit{effective} throughput is defined as the product of the absolute throughput of stage and the accumulated filtration rates.
Note that since we measure the throughput from the pipelined system, the effective throughput of a stage cannot be larger than that of the previous stage.
The results suggest that different datasets experience bottleneck at different stages.
The datasets that attain lower decode filtration rate than the others (i.e., \bench{archie}, \bench{shinjuku}, and \bench{taipei}) are still bottlenecked at the decoder, while the other two datasets are bounded by the DNN object detector.
We observe that the inference of \blobnet never becomes a bottleneck and always matches the throughput of the preceding partial decoding stage.

\subsection{Accuracy Implication of \cova}
Table~\ref{tab:query-result} reports the accuracy results of evaluated queries.
For the BP query, \cova achieves on average \bench{87.3\%} accuracy.
For the CNT query, \cova experiences absolute errors from \bench{0.04} (\bench{archie}) to \bench{1.10} (\bench{taipei}), respectively.
For spatial queries (LBP and LCNT), we do not observe a noticeable difference in accuracy with the temporal queries.
The lack of difference is intuitive since \cova processes the spatial queries by simply restricting the focus of analysis to a certain region of frames.
Therefore, the results of spatial variants are merely a subset of temporal query results.

\begin{table}
       \small
       \caption{Accuracy results of the four evaluated queries for the video datasets. The acronyms for accuracy metric are specified below.}
       \label{tab:query-result}
       \centering
       \resizebox{1.0\linewidth}{!}{
       \begin{tabular}{C{1.4cm}C{1.0cm}C{0.9cm}C{0.9cm}C{0.9cm}C{0.9cm}}
				\hlinewd{0.7pt}
				\multirow{2}{*}{\textbf{Dataset}} & \multirow{2}{*}{\textbf{Object}} & \textbf{BP} & \textbf{CNT} & \textbf{LBP} & \textbf{LCNT} \\
                 &  & (ACC) & (AE) & (ACC) & (AE) \\
				\hlinewd{0.7pt}
				amsterdam & Car & 85.79 & 0.15 & 81.61 & 0.09  \\				
				\hline
				archie & Bus & 86.96 & 0.04 & 90.06 & 0.01 \\				
				\hline
				jackson & Car & 86.13 & 0.10 & 92.01 & 0.05 \\				
				\hline
				shinjuku & Car & 90.15 & 0.30 & 91.31 & 0.05  \\				
				\hline
				taipei & Car & 87.74 & 1.10 & 83.98 & 0.37 \\				
               	\hlinewd{0.7pt}
				average & - & 87.34 & N/A & 87.69 & N/A \\				
               	\hlinewd{0.7pt}
				\multicolumn{6}{r}{* ACC: Accuracy (\textit{\%}), AE: Absolute Error} \\
       \end{tabular}
       }
\end{table}

The results show that the approximate nature of compressed domain analysis introduces accuracy loss.
However, we argue that such modest level of accuracy degradation (10$\sim$20\%) is tolerable to retrospective video analytics, which aims to process large corpus of video data interactively at query time.
The video analytics also inherently produce approximate results due to the nature of noisy analog video data and predictive object detection models.
Moreover, our accuracy results are conservatively calculated by treating the YOLOv4 detection results as ground truth and marking the \cova results as error if they do not match.
However, we empirically observe that there are many cases where YOLOv4 misses small objects when the objects are faraway from the shooting point, while \cova can correctly detect them by successfully tracking blobs even for the small objects and propagating the correct labels to the tracks.
In this case, the correct results of \cova would be marked as false positives due to the erroneous ground truth.

\niparagraph{Discussion.}
As discussed above, approximation is fundamentally inevitable for video analytics, because even the best effort results are still imperfect.
Thus, our goal in designing \cova is to achieve \emph{acceptable} approximation accuracy loss for video analytics.
According to a study~\cite{axgames}, the level of acceptable approximation accuracy loss is higher when the users consider contexts such as application purpose and cost.
We believe that \cova could be a useful tool where analysts can quickly and cost-efficiently extract high-level insights from a large corpus of videos.
For instance, consider an application that monitors traffic in a harbor in Amsterdam (see Table 2).
For binary predicate query, it suffers from 15\% accuracy loss.
However, \cova does not miss the cars completely from the video in most cases since the cars stay in the video for at least several tens of frames (only 2\% of cars are eventually missed).
Hence, if analysts merely wanted to estimate traffic, \cova would be able to offer sufficiently precise results.
We also believe that if an application requires more accurate results, \cova could serve as an initial scanning tool that quickly identifies ``worth-to-be-further-analyzed’’ video clips.

\subsection{Sensitivity Study}

\begin{table}
       \small
       \caption{Raw throughput of four different video codecs on the \bench{libavcodec} and \bench{NVDEC} decoders.}
       \label{tab:codec-sensitivity}
       \centering
       \resizebox{0.96\linewidth}{!}{
       \begin{tabular}{C{1.2cm}C{1.2cm}C{1.7cm}C{2.6cm}}
				\hlinewd{0.7pt}
				\multirow{2}{*}{\textbf{Codec}} & \multicolumn{2}{c}{\textbf{Full Decoding (FPS)}} & \textbf{Partial Decoding} \\
				 & \textbf{NVDEC} & \textbf{libavcodec} & \textbf{(FPS)} \\
				\hlinewd{0.7pt}
				VP8 & 1,590 & 1,802 & 32,774 \\
				\hline
				\textbf{H.264} & \textbf{1,431} & \textbf{1,230} & \textbf{16,761} \\
				\hline
				VP9 & 3,249 & 1,179 & 35,349 \\
				\hline
				H.265 & 3,888 & 2,026 & 25,862 \\
               	\hlinewd{0.7pt}
       \end{tabular}
       }
\end{table}
 \niparagraph{Implication of video codecs.}
We implement the \cova system based on H.264, one of the most widely used video codecs.
However, to demonstrate applicability of \cova to other block-based compression standards, we take three alternatives, VP8, VP9, and HEVC (i.e., H.265), and develop metadata extraction in their partial decoding implementations.
Table~\ref{tab:codec-sensitivity} reports throughput results when using the four different codecs with 720p videos and 32 cores.
The throughput of \textsf{NVDEC} for the four codecs ranges from \bench{1,431 FPS} to \bench{3,888 FPS}, which is significantly lower than the effective throughput of existing cascade systems and thus our problem statement regarding decoding bottleneck still holds true.
In addition, we observe that for all codecs, the full decoding throughput in both software and hardware significantly falls short of throughput of the partial decoding.
This throughput gap allows \cova to construct a cascade architecture that enables blob tracking to run at a higher throughput than the vanilla decoder and effectively lowers the full decode workload.

\begin{figure}
    \centering
    \includegraphics[width=0.95\linewidth]{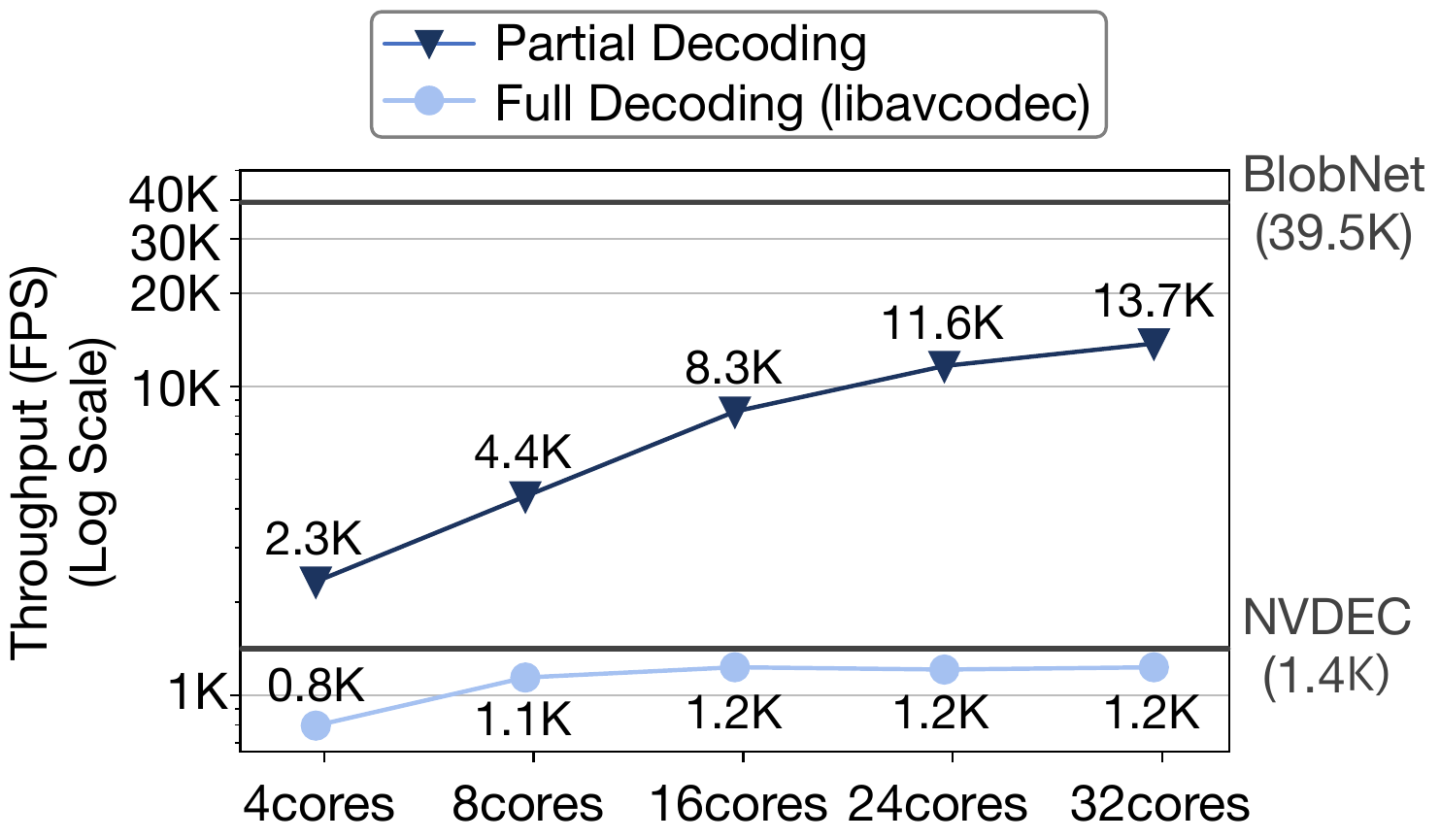}
    \caption{Throughput of partial and full video decoding (libavcodec) on CPU, as the number of cores changes from 4 to 32. For comparison, we also report the throughput of \blobnet and NVDEC, while they have constant throughput since they run on GPU.}
    \label{fig:thread-sensitivity}
\end{figure}
\niparagraph{Implication of CPU parallelism.}
To further analyze the scalability of our parallelization scheme, Figure~\ref{fig:thread-sensitivity} compares the throughput of partial and full decoding as we parallelize them using the varied number of cores from 4 to 32.
We also show the throughput of \blobnet and \textsf{NVDEC} for comparison.
Note that these results are averaged across the datasets.
The results show that the parallelized partial decoder not only scales significantly better than the full decoder when using the same number of cores (i.e., 1.5$\times$ vs. 5.9$\times$), but also largely outperforms the throughput of \textsf{NVDEC}.
Currently, we use all the available cores (32) for partial decoding to optimize for throughput.
However, one may be able to revise the objective function such that it also takes into account resource utilization and energy efficiency, which we leave as a future work.
  \section{Related Work}
\label{sec:related}

A growing body of literature \cite{noscope,probabilistic-predicates,tahoma,adaptive-cascading:cvpr:2017,blazeit,thia:arxiv:2021,video-monitoring-queries,tasm,vstore,smol,vss,focus:osdi:2018,boggart,poms:2018:scanner,videostorm:nsdi:2017,chameleon,videochef,diva} aims to address the computational challenges in video analytics.
\cova differentiates itself by addressing video decoding bottleneck, exploiting compressed-domain analysis.
Further, \cova does not require pre-processing, transcoding, or profiling to obtain the benefits.

\niparagraph{Cascade architectures for binary predicate queries.}
NoScope~\cite{noscope} and Lu et al.~\cite{probabilistic-predicates} use a series of approximate pixel-domain filtering stages to build their cascade.
Tahoma~\cite{tahoma} and Shen et at.~\cite{adaptive-cascading:cvpr:2017} use multiple pipelined  neural networks to build their cascade architecture.
BlazeIt~\cite{blazeit} builds on top of NoScope to support Aggregate and Limit Queries.
All five works aim to increase the effective throughput of the system for raw video frames by filtering a majority of the frames using pixel-domain operators.
Alternatively, Thia~\cite{thia:arxiv:2021} splits up the DNN-inference model using exit points for early termination, similar to the stages of cascade architecture.
In contrast, \xvdec splits the cascade computation between compressed domain and pixel domain to alleviate the decoding bottleneck.

\niparagraph{Spatial queries for video analytics.}
An emerging class of video analytics systems aim to enable queries based on spatial relationship between labeled objects.
Koudas et al.~\cite{video-monitoring-queries} accelerate spatial queries using separate stages for inexpensive DNN-based classification followed by expensive DNN-based object detection.
TASM~\cite{tasm} dynamically adapts the layout of tiles, which partition compressed video frames, based on the spatial location of objects to improve performance.
Unlike the above works, \cova uses compressed domain blob tracking to accelerate spatial queries.
Unlike TASM, \cova does not need to update the compression to gain performance benefits.

\niparagraph{Storage-accuracy trade-off for decoding bottleneck.}
VStore~\cite{vstore} uses a search space of fidelity and encoding/decoding knobs (frame sampling rate, resolution, etc) to optimize for query performance and storage cost.
SMOL~\cite{smol} jointly optimizes complexity of the reference DNN for inference and the resolution of data (360p, 720p, etc), for accuracy-performance trade-off.
VSS~\cite{vss} proposes optimizations for video storage to yield higher read rates and compression ratios.
\cova takes an orthogonal approach of performing approximate blob tracking using compression metadata at \emph{query time}.
Nevertheless, \cova is complementary to the above approaches.

\niparagraph{Ingest time analysis.}
Focus~\cite{focus:osdi:2018} generates approximate labels using an inexpensive DNN and Boggart~\cite{boggart} tracks objects at ingest time to generate additional metadata.
At query time, both Focus and Boggart use the stored metadata to yield improved performance.
Scanner~\cite{poms:2018:scanner} identifies sampling frames offline for pixel domain analysis and skips decoding for all other frames.
In contrast, \xvdec does not require additional metadata and can operate on standard video compression formats.
VideoStorm~\cite{videostorm:nsdi:2017} uses offline profiling data for dynamic load balancing and Chameleon~\cite{chameleon} uses inexpensive online profiling to improve accuracy-resource tradeoff at query time.
These two profiling approaches are orthogonal and complementary to compressed-domain query processing in \xvdec.

\niparagraph{Compressed domain object detection.}
Many prior works~\cite{sukmarg:2000, porikli:2009, laumer:2016, nguyen2020toward} have proposed object detection from compression metadata using classical approaches such as predefined kernels~\cite{poppe:2009} and statistical models~\cite{bombardelli:2018, khatoonabadi:2012, alizadeh:2019}.
Further, the prior works impose restrictions on the compression-time parameters (e.g., 4 frames per GoP), which limit their applicability~\cite{poppe:2009, laumer:2016, alizadeh:2019, nguyen2020toward}.
Liu et al.~\cite{liu:2019}, Wang et al.~\cite{mmnet}, and Wu et al.~\cite{coviar} employ DNNs to detect moving objects using \emph{both} pixel and compressed domain data, training a single model for all datasets.
\blobnet differs from prior works in the following aspects:
(1) \blobnet does not require any pixel data;
(2) \blobnet  does not impose restrictions on the compression parameters;
and (3) \blobnet is trained for given video to compensate the accuracy.

 \section{Conclusion}
\label{sec:conclusion}

Existing cascade systems for video analytics assume to pay significant compute and storage cost for addressing the decoding bottleneck.
Further, the systems are specialized for temporal query to achieve otherwise-unachievable throughput.
To tackle the two limitations, this paper proposes \cova, which splits cascade computation between compressed and uncompressed pixel domain.
Leveraging the unique characteristics of video analytics and video compression algorithm, \cova effectively unclogs the decoding bottleneck while additionally supporting spatial queries.
Our experiments demonstrate that \cova reduces the decoding workload by \bench{{83.6}\%} and offers  \bench{4.8$\times$} system speedup compared to state-of-the-art query-time retrospective video analytics systems, while compromising modest accuracy.

\section{Acknowledgements}
We thank the anonymous reviewers and our shepherd for their comments and feedback.
This work was supported by National Research Foundation of Korea (NRF-2020R1A2C1103088) and Information Technology Research Center (ITRC) support program (IITP-2022-2020-0-01795), both of which are funded by the Ministry of Science and ICT, Korea.
This work was also partly supported by Samsung Electronics Co., Ltd.

\nocite{recsys}
 
\bibliographystyle{unsrt}
\bibliography{paper}

\begin{thebibliography}{10}

\bibitem{cisco-forecast}
Mark Nowell.
\newblock {Cisco VNI Forecast update}.
\newblock
  \url{https://www.ieee802.org/3/ad_hoc/bwa2/public/calls/19_0624/nowell_bwa_01_190624.pdf},
  2021.

\bibitem{noscope}
Daniel Kang, John Emmons, Firas Abuzaid, Peter Bailis, and Matei Zaharia.
\newblock {NoScope: Optimizing Neural Network Queries over Video at Scale}.
\newblock In {\em PVLDB}, 2017.

\bibitem{tahoma}
Michael~R Anderson, Michael Cafarella, German Ros, and Thomas~F Wenisch.
\newblock {Physical Representation-Based Predicate Optimization for a Visual
  Analytics Database}.
\newblock In {\em ICDE}, 2019.

\bibitem{vstore}
Tiantu Xu, Luis~Materon Botelho, and Felix~Xiaozhu Lin.
\newblock {VStore: A Data Store for Analytics on Large Videos}.
\newblock In {\em EuroSys}, 2019.

\bibitem{probabilistic-predicates}
Yao Lu, Aakanksha Chowdhery, Srikanth Kandula, , and Surajit Chaudhuri.
\newblock {Accelerating Machine Learning Inference with Probabilistic
  Predicates}.
\newblock In {\em SIGMOD}, 2018.

\bibitem{smol}
Daniel Kang, Ankit Mathur, Teja Veeramacheneni, Peter Bailis, and Matei
  Zaharia.
\newblock {Jointly Optimizing Preprocessing and Inference for DNN-Based Visual
  Analytics}.
\newblock In {\em PVLDB}, 2020.

\bibitem{video-monitoring-queries}
Nick Koudas, Raymond Li, and Ioannis Xarchakos.
\newblock {Video Monitoring Queries}.
\newblock In {\em ICDE}, 2020.

\bibitem{blazeit}
Daniel Kang, Peter Bailis, and Matei Zaharia.
\newblock {BlazeIt: Optimizing Declarative Aggregation and Limit Queries for
  Neural Network-Based Video Analytics}.
\newblock In {\em PVLDB}, 2019.

\bibitem{focus:osdi:2018}
Kevin Hsieh, Ganesh Ananthanarayanan, Peter Bodik, Shivaram Venkataraman,
  Paramvir Bahl, Matthai Philipose, Phillip~B. Gibbons, and Onur Mutlu.
\newblock {Focus: Querying Large Video Datasets with Low Latency and Low Cost}.
\newblock In {\em OSDI}, 2018.

\bibitem{svq}
Ioannis Xarchakos and Nick Koudas.
\newblock {SVQ: Streaming Video Queries}.
\newblock In {\em SIGMOD}, 2019.

\bibitem{deluceva}
Jingjing Wang and Magdalena Balazinska.
\newblock {Deluceva: Delta-Based Neural Network Inference for Fast Video
  Analytics}.
\newblock In {\em SSDBM}, 2020.

\bibitem{panorama}
Yuhao Zhang and Arun Kumar.
\newblock {Panorama: A Data System for Unbounded Vocabulary Querying over
  Video}.
\newblock In {\em PVLDB}, 2020.

\bibitem{vaas}
Favyen Bastan, Oscar Moll, and Sam Madden.
\newblock {Vaas: Video Analytics At Scale}.
\newblock In {\em PVLDB}, 2020.

\bibitem{chameleon}
Junchen Jiang, Ganesh Ananthanarayanan, Peter Bodík, Siddhartha Sen, and Ion
  Stoica.
\newblock {Chameleon: Scalable Adaptation of Video Analytics}.
\newblock In {\em SIGCOMM}, 2018.

\bibitem{tasm}
Maureen Daum, Brandon Haynes, Dong He, Amrita Mazumdar, and Magdalena
  Balazinska.
\newblock {TASM: A Tile-Based Storage Manager for Video Analytics}.
\newblock In {\em ICDE}, 2021.

\bibitem{nvidia-deepstream}
NVIDIA.
\newblock {DeepStream SDK}.
\newblock \url{https://developer.nvidia.com/deepstream-sdk}, 2021.

\bibitem{nvidia-nvdec}
NVIDIA.
\newblock {Video Codec SDK}.
\newblock \url{https://developer.nvidia.com/nvidia-video-codec-sdk}, 2021.

\bibitem{wiegand:2003:overview-of-h264}
Thomas Wiegand, Gary~J Sullivan, Gisle Bjontegaard, and Ajay Luthra.
\newblock {Overview of the H.264/AVC Video Coding Standard}.
\newblock {\em TCSVT}, 2003.

\bibitem{bewley:2016:sort}
Alex Bewley, Zongyuan Ge, Lionel Ott, Fabio Ramos, and Ben Upcroft.
\newblock Simple online and realtime tracking.
\newblock In {\em ICIP}, 2016.

\bibitem{alizadeh:2019}
Mohammadsadegh Alizadeh and Mohammad Sharifkhani.
\newblock {Compressed Domain Moving Object Detection Based on CRF}.
\newblock {\em TCSVT}, 2020.

\bibitem{zeng:2005}
Wei Zeng, Jun Du, Wen Gao, and Qingming Huang.
\newblock {Robust Moving Object Segmentation on H.264/AVC Compressed Video
  Using the Block-Based MRF Model}.
\newblock {\em Real-Time Imaging}, 2005.

\bibitem{babu:2004:cd-vos}
R.~Babu, Kalpathi Ramakrishnan, and S.H. Srinivasan.
\newblock {Video Object Segmentation: A Compressed Domain Approach}.
\newblock {\em TCSVT}, 2004.

\bibitem{laumer:2016}
Marcus Laumer, Peter Amon, Andreas Hutter, and Andr{\'e} Kaup.
\newblock {Moving Object Detection in the H.264/AVC Compressed Domain}.
\newblock {\em APSIPA}, 2016.

\bibitem{poppe:2009}
Chris Poppe, Sarah~De Bruyne, Tom Paridaens, Peter Lambert, and Rik~Van
  de~Walle.
\newblock {Moving Object Detection in the H.264/AVC Compressed Domain for Video
  Surveillance Applications}.
\newblock {\em Journal of Visual Communication and Image Representation}, 2009.

\bibitem{nguyen2020toward}
Dien~Van Nguyen and Jaehyuk Choi.
\newblock {Toward Scalable Video Analytics Using Compressed-Domain Features at
  the Edge}.
\newblock {\em Applied Sciences}, 2020.

\bibitem{coviar}
Chao-Yuan Wu, Manzil Zaheer, Hexiang Hu, R~Manmatha, Alexander~J Smola, and
  Philipp Kr{\"a}henb{\"u}hl.
\newblock {Compressed Video Action Recognition}.
\newblock In {\em CVPR}, 2018.

\bibitem{mmnet}
Shiyao Wang, Hongchao Lu, Pavel Dmitriev, and Zhidong Deng.
\newblock {Fast Object Detection in Compressed Video}.
\newblock In {\em ICCV}, 2019.

\bibitem{tempunet}
Radu Sibechi, Olaf Booij, Nora Baka, and Peter Bloem.
\newblock {Exploiting Temporality for Semi-Supervised Video Segmentation}.
\newblock In {\em ICCV}, 2019.

\bibitem{adaptive-cascading:cvpr:2017}
Haichen Shen, Seungyeop Han, Matthai Philipose, and Arvind Krishnamurthy.
\newblock {Fast Video Classification via Adaptive Cascading of Deep Models}.
\newblock In {\em CVPR}, 2017.

\bibitem{bae:2014:robust-mot-tracklet}
Seung-Hwan Bae and Kuk-Jin Yoon.
\newblock {Robust Online Multi-Object Tracking Based on Tracklet Confidence and
  Online Discriminative Appearance Learning}.
\newblock In {\em CVPR}, 2014.

\bibitem{yang:2016:temporal}
Min Yang and Yunde Jia.
\newblock {Temporal Dynamic Appearance Modeling for Online Multi-Person
  Tracking}.
\newblock {\em CVIU}, 2016.

\bibitem{xiang:2015:learning-to-track}
Yu~Xiang, Alexandre Alahi, and Silvio Savarese.
\newblock {Learning to Track: Online Multi-Object Tracking by Decision Making}.
\newblock In {\em ICCV}, 2015.

\bibitem{bewley:2014:online-multi-instance-segmentation}
Alex Bewley, Vitor Guizilini, Fabio Ramos, and Ben Upcroft.
\newblock {Online Self-Supervised Multi-Instance Segmentation of Dynamic
  Objects}.
\newblock In {\em ICRA}, 2014.

\bibitem{choi:2015:near-online-multi-target-tracking}
Wongun Choi.
\newblock {Near-Online Multi-Target Tracking with Aggregated Local Flow
  Descriptor}.
\newblock In {\em ICCV}, 2015.

\bibitem{yoon:2015:bayesian-multi-object-tracking}
Ju~Hong Yoon, Ming-Hsuan Yang, Jongwoo Lim, and Kuk-Jin Yoon.
\newblock {Bayesian Multi-Object Tracking Using Motion Context from Multiple
  Objects}.
\newblock In {\em WACV}, 2015.

\bibitem{bewley:2016:alextrac}
Alex Bewley, Lionel Ott, Fabio Ramos, and Ben Upcroft.
\newblock {Alextrac: Affinity Learning by Exploring Temporal Reinforcement
  within Association Chains}.
\newblock In {\em ICRA}, 2016.

\bibitem{nvinfer}
NVIDIA.
\newblock Gst-nvinfer.
\newblock
  \url{https://docs.nvidia.com/metropolis/deepstream/dev-guide/text/DS_plugin_gst-nvinfer.html},
  2021.

\bibitem{dataset-amsterdam}
Webcam Lemmer.
\newblock Binnenhaven lemmer, youtube.
\newblock \url{https://www.youtube.com/watch?v=NyzxJMWxDeo}, 2019.

\bibitem{dataset-jackson}
See~Jackson Hole.
\newblock Jackson hole wyoming usa town square live cam, youtube.
\newblock \url{https://www.youtube.com/watch?v=1EiC9bvVGnk}, 2018.

\bibitem{dataset-shinjuku}
KABUKICHO.
\newblock Shinjuku kabukicho, youtube.
\newblock \url{https://www.youtube.com/watch?v=EHkMjfMw7oU}, 2020.

\bibitem{dataset-taipei}
StarDot Technologies.
\newblock Taiwan new taipei city, youtube.
\newblock \url{https://www.youtube.com/watch?v=INR-B7FwhS8}, 2020.

\bibitem{li:2020:reducto}
Yuanqi Li, Arthi Padmanabhan, Pengzhan Zhao, Yufei Wang, Guoqing~Harry Xu, and
  Ravi Netravali.
\newblock {Reducto: On-Camera Filtering for Resource-Efficient Real-Time Video
  Analytics}.
\newblock In {\em SIGCOMM}, 2020.

\bibitem{traffic-monitoring-1}
M.~Kilger.
\newblock {A Shadow Handler in a Video-Based Real-Time Traffic Monitoring
  System}.
\newblock In {\em WACV}, 1992.

\bibitem{traffic-monitoring-2}
Kostia Robert.
\newblock {Video-Based Traffic Monitoring at Day and Night Vehicle Features
  Detection Tracking}.
\newblock In {\em ITSC}, 2009.

\bibitem{traffic-monitoring-3}
Tariq Abdullah, Ashiq Anjum, M.~Fahim Tariq, Yusuf Baltaci, and Nikos
  Antonopoulos.
\newblock {Traffic Monitoring Using Video Analytics in Clouds}.
\newblock In {\em UCC}, 2014.

\bibitem{video-security}
L.~Snidaro, C.~Micheloni, and C.~Chiavedale.
\newblock {Video Security for Ambient Intelligence}.
\newblock {\em SMC}, 2005.

\bibitem{video-security2}
Minghu Wu, Xiang Li, Cong Liu, Min Liu, Nan Zhao, Juan Wang, Xiangkui Wan,
  Zheheng Rao, and Li~Zhu.
\newblock {Robust Global Motion Estimation for Video Security Based on Improved
  K-Means Clustering}.
\newblock {\em JAIHC}, 2019.

\bibitem{evolution-video-surveillance}
Niels Haering, P\'eter~L. Venetianer, and Alan Lipton.
\newblock {The Evolution of Video Surveillance: An Overview}.
\newblock {\em MVA}, 2008.

\bibitem{healthcare}
P.~Chung, Yung-Ming Kuo, Chin-De Liu, and Chun-Rong Huang.
\newblock {Video Analysis Boosts Healthcare Efficiency and Safety}.
\newblock {\em Spie Newsroom}, 2011.

\bibitem{axgames}
Jongse Park, Emmanuel Amaro, Divya Mahajan, Bradley Thwaites, and Hadi
  Esmaeilzadeh.
\newblock {AxGames: Towards Crowdsourcing Quality Target Determination in
  Approximate Computing}.
\newblock In {\em ASPLOS}, 2016.

\bibitem{thia:arxiv:2021}
Jiashen Cao, Ramyad Hadidi, Joy Arulraj, and Hyesoon Kim.
\newblock {THIA: Accelerating Video Analytics using Early Inference and
  Fine-Grained Query Planning}.
\newblock {\em arXiv}, 2021.

\bibitem{vss}
Brandon Haynes, Maureen Daum, Dong He, Amrita Mazumdar, Magdalena Balazinska,
  Alvin Cheung, and Luis Ceze.
\newblock {VSS: A Storage System for Video Analytics}.
\newblock In {\em SIGMOD}, 2021.

\bibitem{boggart}
Neil Agarwal and Ravi Netravali.
\newblock {Boggart: Accelerating Retrospective Video Analytics via
  Model-Agnostic Ingest Processing}.
\newblock In {\em arXiv}, 2021.

\bibitem{poms:2018:scanner}
Alex Poms, Will Crichton, Pat Hanrahan, and Kayvon Fatahalian.
\newblock {Scanner: Efficient Video Analysis at Scale}.
\newblock {\em TOG}, 2018.

\bibitem{videostorm:nsdi:2017}
Haoyu Zhang, Ganesh Ananthanarayanan, Peter Bodik, Matthai Philipose, Paramvir
  Bahl, and Michael~J. Freedman.
\newblock {Live Video Analytics at Scale with Approximation and
  Delay-Tolerance}.
\newblock In {\em NSDI}, 2017.

\bibitem{videochef}
Ran Xu, Jinkyu Koo, Rakesh Kumar, Peter Bai, Subrata Mitra, Sasa Misailovic,
  and Saurabh Bagchi.
\newblock {VideoChef: Efficient Approximation for Streaming Video Processing
  Pipelines}.
\newblock In {\em ATC}, 2018.

\bibitem{diva}
Mengwei Xu, Tiantu Xu, Yunxin Liu, and Felix~Xiaozhu Lin.
\newblock {Video Analytics with Zero-streaming Cameras}.
\newblock In {\em ATC}, 2021.

\bibitem{sukmarg:2000}
Orachat Sukmarg and Kamisetty~R Rao.
\newblock {Fast Object Detection and Segmentation in MPEG Compressed Domain}.
\newblock In {\em TENCON}, 2000.

\bibitem{porikli:2009}
Fatih Porikli, Faisal Bashir, and Huifang Sun.
\newblock {Compressed Domain Video Object Segmentation}.
\newblock {\em TCSVT}, 2009.

\bibitem{bombardelli:2018}
Fernando Bombardelli, Serhan G{\"u}l, Daniel Becker, Matthias Schmidt, and
  Cornelius Hellge.
\newblock {Efficient Object Tracking in Compressed Video Streams with Graph
  Cuts}.
\newblock In {\em MMSP}, 2018.

\bibitem{khatoonabadi:2012}
Sayed~Hossein Khatoonabadi and Ivan~V. Bajic.
\newblock {Video Object Tracking in the Compressed Domain Using Spatio-Temporal
  Markov Random Fields}.
\newblock {\em TIP}, 2013.

\bibitem{liu:2019}
Qiankun Liu, Bin Liu, Yue Wu, Weihai Li, and Nenghai Yu.
\newblock {Real-Time Online Multi-Object Tracking in Compressed Domain}.
\newblock {\em IEEE Access}, 2019.

\bibitem{recsys}
Mark Zhao, Niket Agarwal, Aarti Basant, Bu\u{g}ra Gedik, Satadru Pan, Mustafa
  Ozdal, Rakesh Komuravelli, Jerry Pan, Tianshu Bao, Haowei Lu, Sundaram
  Narayanan, Jack Langman, Kevin Wilfong, Harsha Rastogi, Carole-Jean Wu,
  Christos Kozyrakis, and Parik Pol.
\newblock {Understanding Data Storage and Ingestion for Large-Scale Deep
  Recommendation Model Training: Industrial Product}.
\newblock In {\em ISCA}, 2022.

\end{thebibliography}

\end{document}